\documentclass[10pt,twocolumn,letterpaper]{article}

\usepackage{wacv}  

\usepackage{graphicx}
\usepackage{amsmath}
\usepackage{amssymb}
\usepackage{booktabs}
\usepackage{multirow}
\usepackage{algorithm}
\usepackage{algorithmicx} 
\usepackage{algpseudocode} 
\usepackage{setspace}
\usepackage{bm}
\makeatletter
\let\ESO@isMEMOIR\relax
\let\AtPageUpperLeft\relax

\let\AddToShipoutPicture\relax

\let\AtTextUpperLeft\relax

\let\ESO@HookI\relax
\let\ESO@HookII\relax
\let\ESO@HookIII\relax
\let\ESO@yoffsetI\relax
\let\ESO@yoffsetII\relax
\makeatother

\usepackage{pdfpages}
\usepackage{tikz}
\usepackage[letterpaper, left=0.68in, right=0.9in, top=1in, bottom=1.0in]{geometry}
\newcommand{\ballnumber}[1]{\tikz[baseline=(myanchor.base)] \node[circle,fill=.,inner sep=1pt] (myanchor) {\color{-.}\bfseries\footnotesize #1};}
\usepackage[pagebackref,breaklinks,colorlinks]{hyperref}

\usepackage[capitalize]{cleveref}
\crefname{section}{Sec.}{Secs.}
\Crefname{section}{Section}{Sections}
\Crefname{table}{Table}{Tables}
\crefname{table}{Tab.}{Tabs.}

\def\wacvPaperID{26} 
\def\confName{WACV}
\def\confYear{2026}

\usepackage[symbol]{footmisc}

\begin{document}

\title{MergeSlide: Continual Model Merging and Task-to-Class Prompt-Aligned Inference for Lifelong Learning on Whole Slide Images}

\author{Doanh C. Bui$^{1,}$\textsuperscript{*}, Ba Hung Ngo$^2$, Hoai Luan Pham$^1$, Khang Nguyen$^3$, Ma\"{i} K. Nguyen$^4$, Yasuhiko Nakashima$^1$ \\
$^1$Nara Institute of Science and Technology, Japan \\
$^2$Graduate School of Data Science, Chonnam National University, South Korea \\
$^3$University of Information Technology, Viet Nam National University Ho Chi Minh City, Viet Nam \\
$^4$ETIS (UMR 8051), CY Cergy Paris University, ENSEA, CNRS, France \\
{\tt\small bui.cao\_doanh.bd2@naist.ac.jp \quad ngohung@chonnam.ac.kr \quad pham.luan@is.naist.jp} \\
{\tt\small khangnttm@uit.edu.vn \quad mai.nguyen-verger@cyu.fr \quad nakashim@is.naist.jp}
}

\maketitle

\footnotetext[1]{Corresponding author.}

\begin{abstract}

Lifelong learning on Whole Slide Images (WSIs) aims to train or fine-tune a unified model sequentially on cancer-related tasks, reducing the resources and effort required for data transfer and processing, especially given the gigabyte-scale size of WSIs.
In this paper, we introduce MergeSlide, a simple yet effective framework that treats lifelong learning as a model merging problem by leveraging a vision-language pathology foundation model. When a new task arrives, it is:
\ballnumber{1} defined with class-aware prompts,
\ballnumber{2} fine-tuned for a few epochs using an MLP-free backbone, and
\ballnumber{3} merged into a unified model using an orthogonal continual merging strategy that preserves performance and mitigates catastrophic forgetting.
For inference under the class-incremental learning (CLASS-IL) setting, where task identity is unknown, we introduce Task-to-Class Prompt-aligned (TCP) inference. Specifically, TCP first identifies the most relevant task using task-level prompts and then applies the corresponding class-aware prompts to generate predictions.
To evaluate MergeSlide, we conduct experiments on a stream of six TCGA datasets. The results show that MergeSlide outperforms both rehearsal-based continual learning and vision-language zero-shot baselines. Code and data are available at \url{https://github.com/caodoanh2001/MergeSlide}.

\end{abstract}

\vspace{-0.5cm}
\section{Introduction}
\label{sec:intro}

Whole Slide Images (WSIs) provide a comprehensive view of tissue at the cellular level, which is essential for cancer diagnosis and prognosis using digital computational pathology tools and protocols \cite{wu2023artificial}. In recent years, multiple-instance learning (MIL) models \cite{ilse2018attention,clam,transmil,dtfd} have become the main approach for WSI-related tasks such as tumor classification and cancer subtyping. However, with the rapid adoption of WSIs and the emergence of new cancer-related tasks, developing computational pathology tools has become increasingly challenging \cite{gou2025queryable}. {Due to their gigabyte-scale size, training deep unified multi-purpose models for WSIs is time-consuming, as it often requires substantial resources for data transfer, preprocessing, feature extraction, and model storage \cite{shen2022federated}. Moreover, data privacy concerns across institutions restrict data sharing, hindering the development of multi-purpose models that leverage diverse WSIs from multiple sources.}

\begin{figure}[!http]
\centerline{\includegraphics[width=0.49\textwidth]{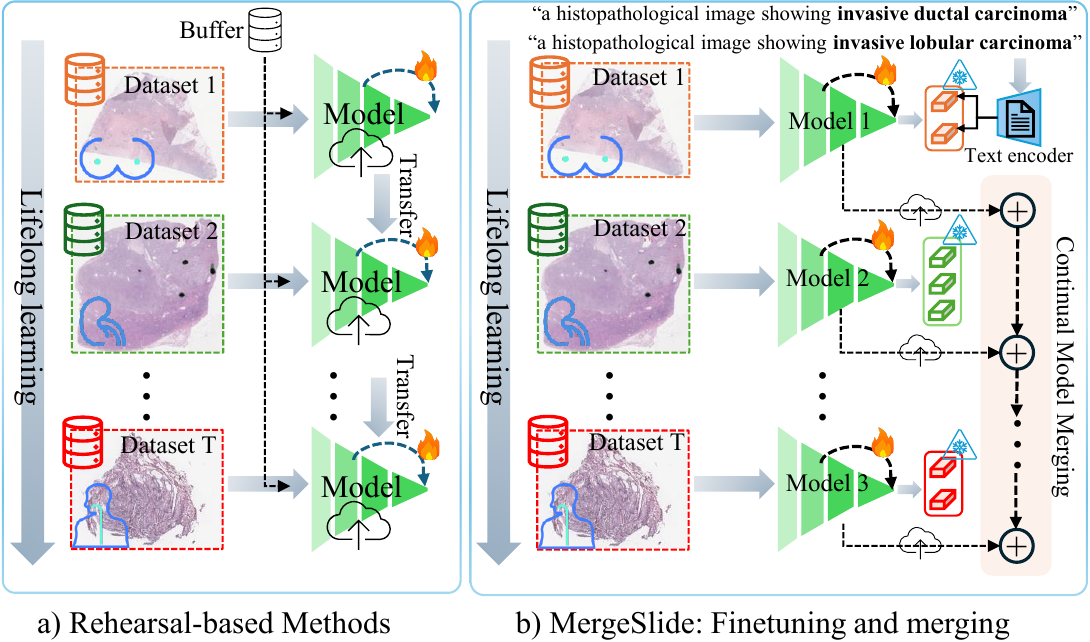}}
\caption{\textbf{(a) Rehearsal-based methods}; \textbf{(b) Task-specific fine-tuning and merging in MergeSlide}; MergeSlide first trains an MLP-free model offline using frozen class-aware prompt embeddings. Then, the weights are merged task-by-task.}
\label{fig:thumbnail} 
\end{figure}

\begin{figure*}[http]
\centerline{\includegraphics[width=0.85\textwidth]{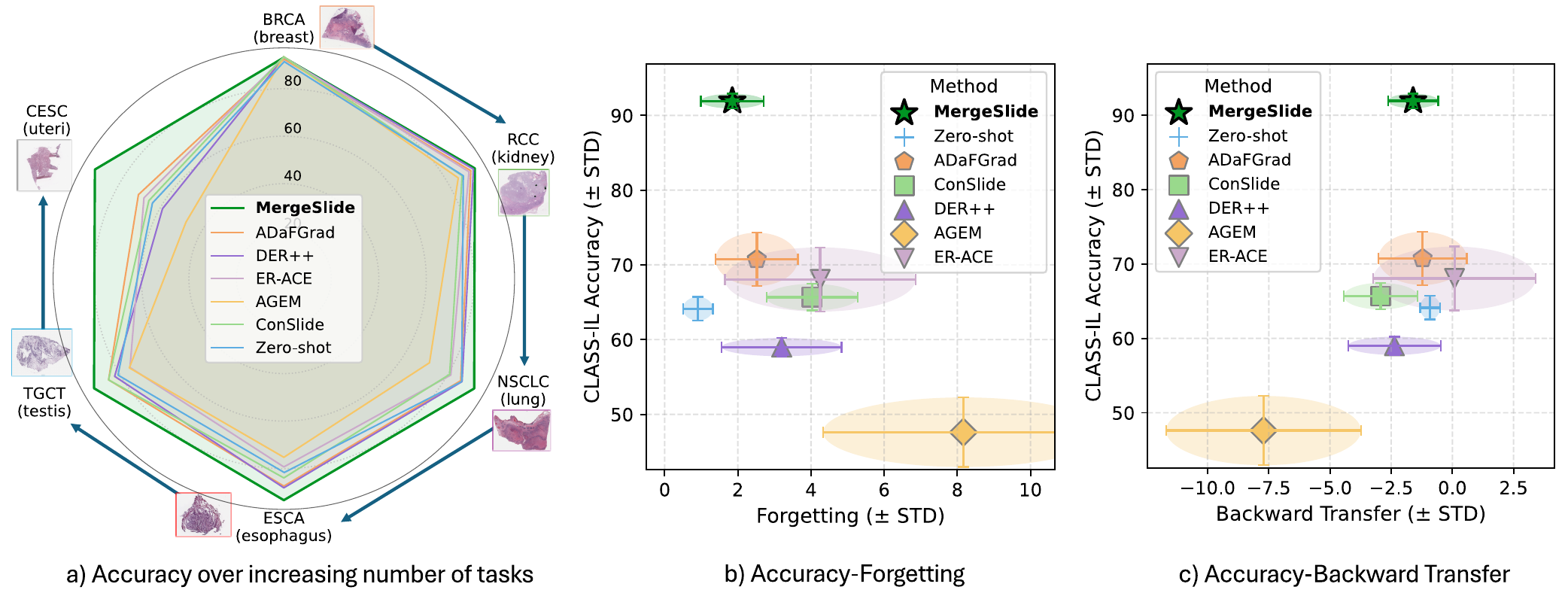}}
\caption{\textbf{Performance comparison of MergeSlide with other continual learning methods.} (a) Accuracy as tasks accumulate. (b) Accuracy vs. Forgetting. (c) Accuracy vs. Backward Transfer, both after the final task.}
\label{fig:radarchart}
\end{figure*}

An effective solution could be to develop a single core model or tool that is continuous, adaptive, and scalable as new tasks are introduced. A naive approach might be fine-tuning a baseline model task-by-task, but this often leads to catastrophic forgetting \cite{mccloskey1989catastrophic,boschini2022transfer}, which is a phenomenon where a model trained on a new task forgets or performs poorly on previously learned tasks. To tackle this, traditional lifelong learning \cite{de2021continual} (also known as continual learning) approaches are possible \cite{huang2023conslide,gou2025queryable}. These methods are generally categorized into regularization-based and rehearsal-based techniques. Regularization-based models \cite{li2017learning,rebuffi2017icarl,kirkpatrick2017overcoming,zenke2017continual,aljundi2018memory} typically incorporate mechanisms to prevent the model from deviating too far from previous tasks when learning new ones. In contrast, rehearsal-based models \cite{agem,derpp,erace,huang2023conslide} relax this constraint by allowing the model to revisit a small buffer containing samples from previous tasks. As an early attempt to apply lifelong learning to WSIs, ConSlide \cite{huang2023conslide} introduced a breakup-and-recognize strategy to diversify buffered WSIs. Later on, with the emergence of pathology vision-language foundation models (VLMs) \cite{lu2023visual,huang2023visual,lu2024visual,ding2024multimodal} that align images with diagnostic texts, \cite{huang2024free} demonstrated that such models can effectively boost downstream WSI performance through vision-text interaction. Building on this, \cite{gou2025queryable} employed class prompts as query prototypes to enhance lifelong learning using foundation models. Nonetheless, we argue that these models have certain drawbacks:
Rehearsal-based methods often outperform others but \textit{\textcircled{\raisebox{-0.9pt}{1}} pose a risk of data leakage as well as additional memories due to the use of buffers.}
\textit{\textcircled{\raisebox{-0.9pt}{2}} The implementation of these models typically assumes that tasks have the same number of classes, {and that the maximum logits for classification must be predefined regardless of the number of incoming tasks}}, which is impractical. In practice, the number of classes should be defined flexibly by domain experts rather than constrained by the method. Finally, \textit{\textcircled{\raisebox{-0.9pt}{3}} these models are sensitive to the order of tasks}; that is, their performance is not guaranteed to remain consistent when the task order changes, as their training schemes heavily rely on interactions with previous parameters or WSIs stored in a buffer. In another line of research, model merging methods \cite{ilharco2022editing,yadav2023ties,tang2024merging} have shown that it is possible to merge parameters from individually trained models across different tasks by leveraging task vectors \cite{ilharco2022editing}, which measure the differences between each task and a well-pretrained base model that captures general knowledge. This direction is promising, as it minimizes or avoids interference with parameters across tasks during training. Nonetheless, its performance relative to continual learning models and its applicability to computational pathology remain underexplored, due to the previous lack of pathology foundation models.

In this paper, we introduce \textbf{MergeSlide}, a framework that enables efficient lifelong learning for whole slide image (WSI) analysis. An overview of MergeSlide and its comparison with rehearsal-based continual learning methods are shown in Fig.~\ref{fig:thumbnail}. Simply put, we treat lifelong learning on WSIs as a model merging process. When a new task emerges, we: \ballnumber{1} \textit{design class-aware prompts to effectively describe the target cancer subtypes and obtain their embeddings using a pathology vision--language model};  
\ballnumber{2} \textit{fine-tune for a few epochs using an MLP-free backbone with randomly sampled patches}, making it less time-consuming to develop a unified, multi-purpose WSI model. The resulting task-specific weights are then  
\ballnumber{3} \textit{incrementally merged into a unified model that retains the ability to handle all previously seen cancer-related tasks}, eliminating the need for a rehearsal buffer.  
For inference, the backbone is used solely for slide embedding, and predictions are obtained via dot-product similarity with task-level and class-aware prompt embeddings. Thanks to this design, {MergeSlide addresses the following disadvantages:  
\textit{\textcircled{\raisebox{-0.9pt}{1}} MergeSlide reduces the time required to develop a unified, multi-purpose giga-scale WSI model:} since fine-tuning is performed individually for each task and then merged into a unified model, it avoids repeatedly training the entire model from scratch when new tasks appear.
\textit{\textcircled{\raisebox{-0.9pt}{2}} MergeSlide preserves data privacy:} it does not require a rehearsal buffer, thereby avoiding potential data leakage.}
Furthermore, one might argue that with the development of pathology VLMs, zero-shot learning on WSIs \cite{mahapatra2021medical,lu2023visual,javed2024cplip} could outperform continual learning approaches simply by defining effective prompt prototypes for new tasks. However, in this study, we demonstrate that \textit{\textcircled{\raisebox{-0.9pt}{3}} MergeSlide, with just a few finetuning epochs on a pre-trained foundation pathology model, significantly outperforms the zero-shot learning approach}, while still preserving both computational efficiency and practical applicability.
Finally, since MergeSlide accumulates fine-tuned models task by task into a single unified model through merging, \textit{\textcircled{\raisebox{-0.9pt}{4}} it is asymptotically order-free}, i.e., nearly independent of task order, as performance variation across task permutations is minimal.
All of these make MergeSlide continuous, adaptive, and efficiently scalable for lifelong learning on WSI-related tasks.
We evaluate MergeSlide on six TCGA cancer subtyping tasks under two settings: class-incremental (CLASS-IL), where task identity is unknown, and task-incremental (TASK-IL), where it is provided. Results show that MergeSlide outperforms continual learning and zero-shot baselines, achieving the best balance between accuracy and forgetting (Fig.~\ref{fig:radarchart}), demonstrating its practical utility. Our study is summarized as follows:

\begin{itemize}
\item We frame lifelong learning on WSIs as a continual model merging problem, where each task is trained separately and then incrementally merged into a unified, multi-purpose model.
\item For each task, we introduce MLP-free Task-Specific Fine-Tuning, using a pre-trained slide aggregator and predefined class prompts for prediction. {This enables open-class generalization, unlike MLP-based methods that must predefine logits regardless of the number of tasks.}
\item We present {Task-to-Class Prompt-Aligned Inference (TCP)} for the CLASS-IL scenario. In this approach, a slide embedding is aligned with class- and task-level prompt embeddings to predict the cancer subtype.
\item MergeSlide outperforms both continual learning and vision-language zero-shot baselines across six TCGA datasets under CLASS-IL and TASK-IL settings.
\end{itemize}

\section{Related Works}
\label{sec:relatedwork}
\noindent\textbf{Lifelong Learning.} Lifelong learning methods can be categorized into regularization-based and rehearsal-based approaches. For example, in regularization-based methods, EWC \cite{kirkpatrick2017overcoming} constrains model updates by penalizing changes to weights deemed important for previous tasks. In contrast, rehearsal-based methods often outperform regularization-based ones. For instance, GDumb \cite{gdumb} mitigates forgetting by retraining on a balanced memory buffer. ER-ACE \cite{erace} adjusts loss weighting during replay to balance learning between old and new tasks. A-GEM \cite{agem} reduces task interference by projecting gradients to preserve knowledge from previous tasks. DER++ \cite{derpp} enhances replay via logit distillation. There is also a line of research based on orthogonal gradient projection (OGP) that attempts to update the gradient of a new task in an orthogonal direction to the subspace spanned by previous tasks, where the task subspace is constructed using Singular Value Decomposition (SVD) \cite{lin2022trgp,yang2025revisiting}. Regarding lifelong learning for computational pathology, ConSlide \cite{huang2023conslide} pioneers continual learning for WSIs by introducing a breakup-reorganize strategy to diversify buffered WSIs. ADaFGrad \cite{adafgrad} distills gradients between current and previous classification weights \cite{du2020agree}, and applies contrastive learning between WSIs and class/negative prompts.


\noindent\textbf{Model Merging Strategies.} Model merging combines weights from task-specific models into a unified model while aiming to preserve performance. It typically relies on task vectors, which are differences between task-specific and general knowledge weights \cite{ilharco2022editing}. Recent works have explored weighting strategies to merge these vectors effectively \cite{ilharco2022editing,yadav2023ties,yang2023adamerging,tang2024merging}, including Mixture-of-Experts approaches \cite{yang2023adamerging,tang2024merging} and continual merging based on task vector orthogonality \cite{tang2025merging}. This strategy suits lifelong learning on WSIs, enabling offline training, preserving data privacy, and reducing cross-institution transfer. 

\noindent\textbf{Pathology Foundation Models.} Recent work in pathology image analysis focuses on foundation models using CNNs \cite{he2016deep} or ViTs \cite{dosovitskiy2020image}, pretrained with self-supervised methods like DINO \cite{caron2021emerging} and MoCov3 \cite{chen2021empirical}. Models such as CTransPath \cite{wang2022transformer}, UNI \cite{chen2024towards} and ProvGigaPath \cite{xu2024whole} use tile- or slide-level image data only, while multimodal models like PLIP \cite{huang2023visual} and CONCH \cite{lu2024visual} leverage large vision–text pairs at the tile level. TITAN \cite{ding2024multimodal} operates at the slide level. These models are trained with vision-language contrastive learning (e.g., iBOT \cite{zhou2021ibot}, CoCa \cite{yu2022coca}). \cite{huang2024free} has shown that that using prompt embeddings from text encoder of CONCH improves downstream WSI classification tasks.

\section{Methodology}
\label{sec:methodology}

\begin{figure*}[!h]
\centerline{\includegraphics[width=0.9\textwidth]{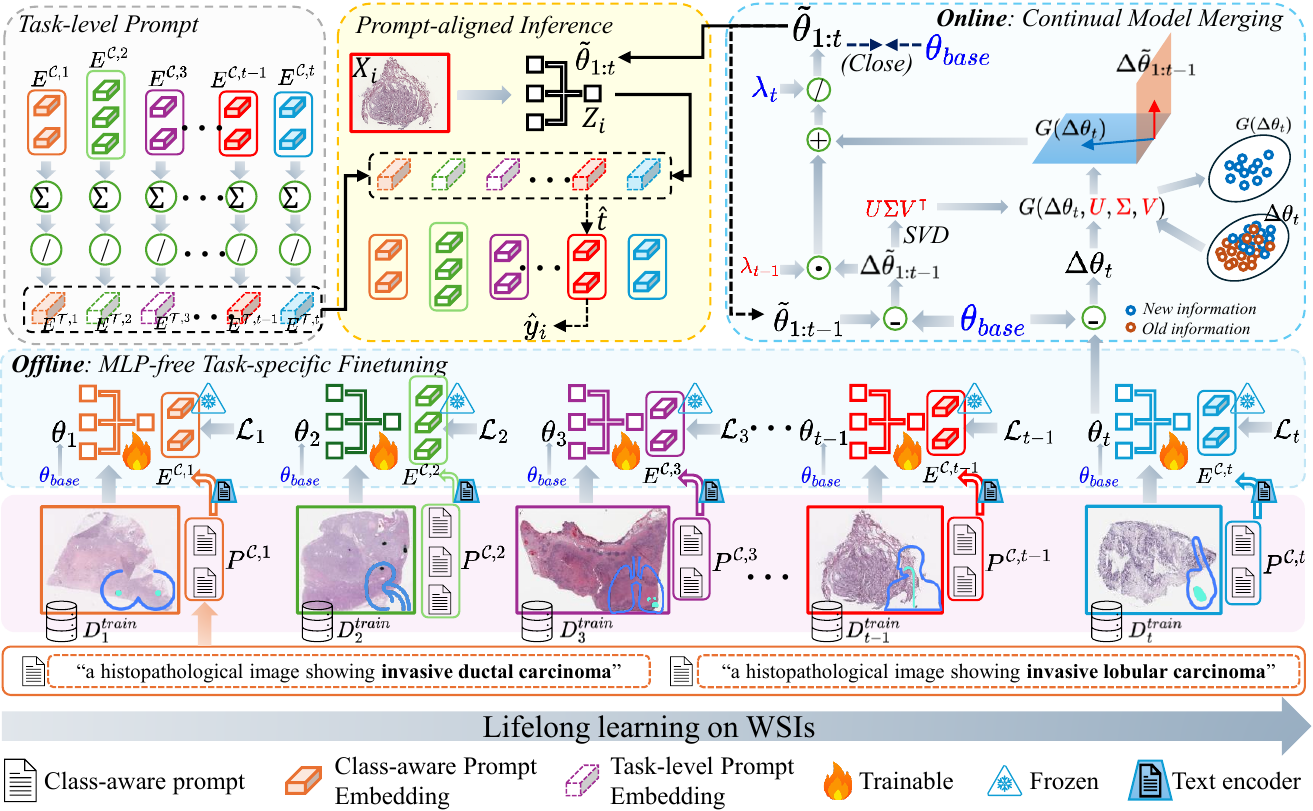}}
\caption{\textbf{Illustration of MergeSlide}. For each new task $t$, class-aware prompt embeddings $E^{\mathcal{C},t}$ are created. An \textbf{MLP-free fine-tuning step} (Sec.~\ref{sec:mlp-free}) initializes the slide aggregator $f_{\mathcal{A}}$ from base weights $\theta_{base}$, yielding task-specific weights $\theta_t$. These are then fused via \textbf{online continual model merging} (Sec.~\ref{sec:cmm}) into unified weights $\tilde{\theta}_{1:t}$ capable of handling all tasks up to $t$. Finally, \textbf{Task-to-class prompt-aligned inference} (Sec.~\ref{sec:pai}) identifies the relevant task using task-level prompts and predicts based on class-aware embeddings.}
\label{fig:mergeslide} 
\end{figure*}

\subsection{Preliminaries}

\noindent\textbf{Problem.} Given a stream of $T$ datasets $\mathcal{D} = \{D_t\}_{t=1}^T$, where each $D_t = \{D^{train}_t, D^{test}_t\}$ contains WSIs from $c_t$ cancer subtypes, the goal is to fine-tune a pre-trained slide aggregator $f_{\mathcal{A}}$ on each $D^{train}_t$ and incrementally merge the weights $\{\theta_t\}_{t=1}^{T}$ into unified weights $\tilde{\theta}_{1:T}$. The final model $f_{\mathcal{A}}(\cdot, \tilde{\theta}_{1:T})$ should generalize well across all tasks, maintaining stable performance on $\{D^{test}_t\}_{t=1}^T$. We consider two settings: (1) CLASS-IL, where task identity is unknown; and (2) TASK-IL, where it is provided at inference.

\noindent\textbf{WSI Processing.} First, for each WSI $X_i \in D_t$, it first goes through a preprocessing step. Following previous studies \cite{huang2023conslide,transmil,dtfd}, we adopt the CLAM pipeline \cite{clam}, which first segments tissue regions in the WSI using Otsu thresholding, then tiles them into patches. Each patch is processed through a frozen feature extractor; we utilize the vision encoder from TITAN \cite{ding2024multimodal}, a pathology VLM, to extract a patch embedding $v_i \in \mathbb{R}^{d_{vis}}$, where $\mathbb{R}^{d_{vis}}$ is the embedding size of each patch embedding. {It is worth noting that TITAN was chosen because it is the state-of-the-art pathology VLM, surpassing other available models such as CONCH \cite{lu2024visual} and PLIP \cite{huang2023visual}. As a direct update of CONCH, TITAN delivers stronger performance, while models like CTransPath \cite{wang2022transformer}, UNI \cite{chen2024towards} and ProvGigaPath \cite{xu2024whole} are excluded due to the absence of a well-aligned text encoder.} Finally, we obtain a bag of patch embeddings $B_i = \{v_i\}^{|B_i|}_{i=1}$.

\subsection{MergeSlide}

\noindent\textbf{Overview.} {Fig.~\ref{fig:mergeslide} illustrates MergeSlide, which consists of three stages: \textit{\textcircled{\raisebox{-0.9pt}{1}} MLP-free Fine-tuning:} For each new $t$-th task, a set of class prompts $P^{\mathcal{C},t} = \{p^{\mathcal{C},t}_i\}_{i=1}^{c_t}$ is defined, and only the slide aggregator ${f}_{\mathcal{A}}(\cdot,\theta_t)$ is fine-tuned. After training $f_{\mathcal{A}}$ on the $t$-th task, we perform \textit{\textcircled{\raisebox{-0.9pt}{2}} Model Merging:} the trained weights $\theta_t$ are merged with the accumulated weights $\tilde{\theta}_{1:t}$, integrating knowledge from all tasks so far. For inference, we introduce \textit{\textcircled{\raisebox{-0.9pt}{3}} Task-to-Class Prompt-aligned Inference:} under the CLASS-IL setting, where task identity is unknown. Given a bag of patch embeddings $B_i$ from a WSI, $B_i$ is subsampled into $B'_i$ to reduce patch count. \textit{MergeSlide} then compares the slide-level embedding $Z_i =f_{\mathcal{A}}(B'_i,\tilde{\theta}_{1:t})$ with task prompts $\mathcal{P}^\mathcal{T} = \{P^{\mathcal{T},t}\}_{t=1}^{|\mathcal{D}|}$ to predict the most relevant task $\textcolor{orange}{\hat{t}}$, and uses the corresponding class prompts $P^{\mathcal{C}, \textcolor{orange}{\hat{t}}}$ for final cancer subtype prediction. Alg.~1 in the Supplementary Material summarizes the procedure.}

\subsubsection{MLP-free Task-specific Fine-tuning}\label{sec:mlp-free} When the new $t$-th task arrives, we first initialize the weights $\theta_t$ of the slide aggregator $f_{\mathcal{A}}$ from the TITAN foundation model, denoted as $\theta_{{base}}$. A set of class prompts $P^{\mathcal{C},t} = \{p^{\mathcal{C},t}_j\}_{j=1}^{c_t}$ is then defined. The text encoder of TITAN {(which is frozen during the fine-tuning process)} is used to generate embeddings for these prompts, denoted as $E^{\mathcal{C},t} = \{e^{\mathcal{C},t}_j\}_{j=1}^{c_t}$, where $e^{\mathcal{C},t}_j \in \mathbb{R}^{1 \times d_{{text}}}$ and $d_{{text}}$ is the embedding dimension of each prompt.
To minimize training cost, given a bag of patches $B_i \in D_t$ with corresponding label $y_i$, we first apply random sampling to obtain only $K$ patch embeddings, resulting in $B'_i = \{v_j\}_{j=1}^K$ ($B'_i \in B_i$). This sampled bag $B'_i$ is then passed through $f_{\mathcal{A}}$ to obtain a slide-level embedding $Z_i$, which is subsequently compared with $E^{\mathcal{C},t}$ to compute the final logit:
\begin{equation}
Z_i = f_{\mathcal{A}}(B'_i, \theta_t), \quad \hat{p}_i = \{Z_i \cdot ({e^{\mathcal{C},t}_j})^\top| \forall e^{\mathcal{C},t}_j \in E^{\mathcal{C},t}\},
\label{eq:1}
\end{equation}
\noindent where $\hat{p}_i \in \mathbb{R}^{c_t}$ includes predicted logits for the $t$-th task. Initially, the task-specific weights are set as $\theta_t = \theta_{{base}}$. The fine-tuning process simply minimizes the cross-entropy loss for $N_e$ epochs:
\begin{equation}
\min_{\theta_t}\mathcal{L}_t (y_i, \hat{p}_i) =  \min_{\theta_t} \left(-y_i \log \hat{p}_i \right),
\label{eq:2}
\end{equation}

\noindent where $y_i \in \{0,1\}^{c_t}$ is the one-hot encoded ground truth.

\noindent\subsubsection{Orthogonal Continual Model Merging Strategy}\label{sec:cmm} From a continual model merging perspective, once the fine-tuning process for the new task $t$ is completed, its weights are merged into the accumulated weights from previous tasks. In this manner, we consider three sets of parameters: \textcircled{\raisebox{-0.9pt}{1}}: $\theta_{{base}} \in \mathbb{R}^{m\times n}$, a well-pretrained model assumed to capture general pathological knowledge; \textcircled{\raisebox{-0.9pt}{2}}: $\tilde{\theta}_{1:t-1} \in \mathbb{R}^{m\times n}$, the cumulatively merged parameters from tasks 1 to $t{-}1$; and \textcircled{\raisebox{-0.9pt}{3}}: $\theta_t \in \mathbb{R}^{m\times n}$, the newly fine-tuned parameters for the $t$-th task. \textit{The goal is to effectively merge $\theta_t$ into $\tilde{\theta}_{1:t-1}$ to obtain $\tilde{\theta}_{1:t}$, while leveraging $\theta_{{base}}$.} Following \cite{ilharco2022editing,tang2025merging}, we adopt the general update rule as follows:

\begin{equation}
\tilde{\theta}_{1:t} = \theta_{base} + \frac{\lambda_{t-1} \Delta\tilde{\theta}_{1:t-1} + G(\Delta \theta_t,\Delta\tilde{\theta}_{1:t-1})}{\lambda_t},
\label{eq:merge}
\end{equation}

\noindent where $\Delta \theta_t$ denotes the task vector \cite{ilharco2022editing}, defined as the difference between any parameter set $\theta_t$ and the general knowledge $\theta_{base}$: $\Delta \theta_t = \theta_t - \theta_{base}$. $G(\cdot)$ is a projection function designed to map the current task vector $\Delta \theta_t$ into a more efficient space. Finally, $\lambda_{t-1}$ and $\lambda_t$ are used to control the contributions of the previously merged parameters and the new task parameters, respectively. 
To derive an efficient $\tilde{\theta}_{1:t}$, we follow the principle that it should balance between $\theta_{{base}}$ and $\tilde{\theta}_{1:t-1}$: \ballnumber{a} $\tilde{\theta}_{1:t}$ must be sufficiently distant from $\tilde{\theta}_{1:t-1}$ to retain new knowledge (e.g., orthogonality \cite{lin2022trgp,yang2025revisiting}), while \ballnumber{b} ensuring that the new merged model $\tilde{\theta}_{1:t}$ does not drift too far from the general pathology knowledge $\theta_{{base}}$ provided by the foundation model. As shown in Eq.~\eqref{eq:merge}, this trade-off is governed by $\lambda_t$, $\lambda_{t-1}$, and the projection function $G(\cdot)$.

\textit{\textbf{To satisfy Condition \ballnumber{a}}}, drawing inspiration from OGP-based continual learning \cite{lin2022trgp, yang2025revisiting} and model merging techniques \cite{tang2025merging}, we aim to make $\theta_t$ orthogonal to $\tilde{\theta}_{1:t-1}$. First, we decompose $\Delta\tilde{\theta}_{1:t-1} \in \mathbb{R}^{m \times n}$ by computing the full singular value decomposition (SVD) of the previously merged model: $\Delta\tilde{\theta}_{1:t-1} = U\Sigma V^\top$, where $U \in \mathbb{R}^{m \times m}$ and $V \in \mathbb{R}^{n \times n}$ are orthogonal matrices, and $\Sigma \in \mathbb{R}^{m \times n}$ is a diagonal matrix. Since the parameter matrix $\Delta\tilde{\theta}_{1:t-1} \in \mathbb{R}^{m \times n}$ projects an $m$-dimensional vector to an $n$-dimensional one, $U$ and $V$ capture the main directions (patterns) of the input and output spaces, respectively, from previous tasks. The projection $G(\cdot)$ is then defined as:

\begin{equation}
G\left( \Delta \theta_t \right)
= U \left( \left( U^\top \Delta \theta_t V \right) \odot M \right) V^\top,
\label{eq:project}
\end{equation}

\noindent where $M$ is a zero-diagonal masking matrix that filters out the shared components between $\Delta \theta_t$ and $\Delta \tilde{\theta}_{1:t-1}$. Thanks to Eq.~\eqref{eq:project}, $\Delta \theta_t$ is projected onto the space orthogonal to $\Delta \tilde{\theta}_{1:t-1}$, i.e., $\langle G\left( \Delta \theta_t \right),\ \Delta \tilde{\theta}_{1:t-1} \rangle_F = 0$, ensuring their separability. To justify this, consider the product $G\left( \Delta \theta_t \right) \Delta \tilde{\theta}_{1:t-1}$, which simplifies to $\left( \left( U^\top \Delta \theta_t V \right) \odot M \right) \Sigma$. Since $M$ is zero on the diagonal and $\Sigma$ is diagonal, the Frobenius inner product $\langle M \odot \Sigma,\, \cdot \rangle_F = 0$. Consequently, the projection is orthogonal. \textit{In this manner, $G(\Delta\theta_t)$ extracts the novel information of the $t$-th task relative to previous tasks.} Specifically, if $\Delta\theta_t$ is well aligned with $\tilde{\theta}_{1:t}$, i.e., not novel, its projection onto the SVD basis yields dominant diagonal values. These are masked by $M$, resulting in a small $G(\Delta\theta_t)$. In contrast, if the task is novel, the off-diagonal components dominate and are preserved by $M$, producing a larger $G(\Delta\theta_t)$.

\textit{\textbf{To satisfy Condition \ballnumber{b}}}, a consistent $\|\tilde{\theta}_{1:t} - \theta_{base}\|_2$ is maintained as the number of tasks increases, and $\lambda_t$ is designed accordingly as follows:

\begin{equation}
\lambda_t = t \cdot \frac{\left\| \lambda_{t-1} \Delta \tilde{\theta}_{1:t-1} + G(\Delta \theta_t) \right\|_2}{\sum_{i=1}^{t} \left\| \theta_i \right\|_2}, \text{ where } \lambda_1 = 1.
\label{eq:lambda}
\end{equation}

\noindent To further explain how $\|\tilde{\theta}_{1:t} - \theta_{base}\|_2$ becomes consistent, let $U_t = \lambda_{t-1} \Delta\tilde{\theta}_{1:t-1} + G(\Delta \theta_t, \Delta\tilde{\theta}_{1:t-1})$. By substituting Eq. \eqref{eq:lambda} into Eq. \eqref{eq:merge} and taking the norm $\| \tilde{\theta}_{1:t} - \theta_{base} \|_2$, we obtain:

\begin{equation}
\|\tilde{\theta}_{1:t} - \theta_{base}\|_2 = \frac{1}{t} \sum_{i=1}^{t} \left\| \theta_i \right\|_2 \cdot \left\| \frac{U_t}{\|U_t\|_2} \right\|_2 = \frac{1}{t} \sum_{i=1}^{t} \left\| \theta_i \right\|_2.
\label{eq:merge}
\end{equation}

\noindent Obviously, $\frac{1}{t} \sum_{i=1}^{t} \left\| \theta_i \right\|_2 \leq \max_{i \in \{1,\dots,t\}} \left\| \theta_i \right\|_2$. Therefore, with $\lambda_t$ computed as in Eq. \eqref{eq:lambda}, the norm $\|\tilde{\theta}_{1:t} - \theta_{base}\|_2$ is upper bounded by the maximum norm of $\theta_i$. In other words, the merged weights are ensured to never drift farther from the base model than the largest single-task norm observed so far.



\noindent\subsubsection{Task-to-Class Prompt-aligned Inference}\label{sec:pai} After training on all $T$ tasks, we introduce Task-to-Class Prompt-aligned Inference to effectively leverage a pathology vision-language foundation model in the {CLASS-IL scenario}, where the target task is unknown. This process decomposes CLASS-IL scenario into two steps: \textit{\textcircled{\raisebox{-0.9pt}{1}} identifying the task most aligned with the given input WSI}, and \textit{\textcircled{\raisebox{-0.9pt}{2}} obtaining a prediction by comparing the input’s similarity to the class-aware prompts of the identified task}. To identify the most suitable task, we construct a set of $T$ task-level prompt embeddings $\mathcal{E}^\mathcal{T} = \{E^{\mathcal{T},t}\}_{t=1}^T$. For each task prompt $E^{\mathcal{T},t}$, we do not introduce a new prompt; instead, we leverage its corresponding set of class-aware embeddings $E^{\mathcal{C},t}$. Each $E^{\mathcal{C},t}$ contains a set of class-specific embeddings, which we average to obtain the task-level embedding: $E^{\mathcal{T},t} = \frac{1}{c_t} \sum E^{\mathcal{C},t}$. Then, to identify the task most aligned with a given input WSI $X_i$, we compute dot-product similarity between its slide embedding, produced by $f_{\mathcal{A}}(X_i,\tilde{\theta}_{1:T})$ and all task-level prompt embeddings in $\mathcal{E}^\mathcal{T}$. The task prediction $\textcolor{orange}{\hat{t}}$ is determined by the highest similarity score:

\begin{equation}
\textcolor{orange}{\hat{t}} = \arg \max_{t} \left\{Z_i \cdot (E^{\mathcal{T},t})^\top | \forall E^{\mathcal{T},t} \in \mathcal{E}^{\mathcal{T}}\right\}.
\end{equation}

\noindent Once the predicted target task $\textcolor{orange}{\hat{t}}$ is identified, its corresponding class-aware prompt embeddings $E^{\mathcal{C},\textcolor{orange}{\hat{t}}}$ are used to perform cancer subtyping. The prediction $\hat{p}_i$ is obtained by computing the dot-product similarity between the slide embedding $Z_i$ and the set of class-aware prompt embeddings for the $\textcolor{orange}{\hat{t}}$-th task. The predicted sub-cancer subtype is then defined by selecting the class corresponding to the maximum logit:

\begin{equation}
\hat{p}_i = \{Z_i \cdot (e^{\mathcal{C},\textcolor{orange}{\hat{t}}}_{j})^\top\vert\forall e^{\mathcal{C},\textcolor{orange}{\hat{t}}}_{j}\in E^{\mathcal{C},{\textcolor{orange}{\hat{t}}}}\}, \quad \hat{y}_i = \arg\max \hat{p}_i.
\end{equation}

\section{Experiments}
\label{sec:experiments}
\noindent\textbf{Datasets.} We use six TCGA cancer subtyping cohorts from the Genomic Data Commons (GDC) Data Portal\footnote[2]{\url{https://portal.gdc.cancer.gov/}} for training and evaluation: TCGA-BRCA (breast), TCGA-RCC (kidney), TCGA-NSCLC (lung), TCGA-ESCA (esophagus), TCGA-TGCT (testis), and TCGA-CESC (cervix uteri). Each cohort is split into 10 folds, and 10-fold cross-validation is performed to evaluate MergeSlide against other continual learning methods. The number of subtypes per cohort is reported in Tab.~\ref{tab:slide_counts}. Slide counts vary across datasets, with some cohorts having abundant data (\textit{common cohorts}) and others having limited data (\textit{rare cohorts}).

\begin{table}[ht]
\centering
\resizebox{1\linewidth}{!}{\begin{tabular}{llr}
\toprule
\textbf{Dataset} & \textbf{Subtype}  & \textbf{\# WSIs} \\
\midrule
\multirow{2}{*}{TCGA-BRCA (\textcolor{blue}{B})} & invasive ductal carcinoma (IDC) & 726  \\
   & invasive lobular carcinoma (ILC) & 149  \\
\midrule
\multirow{3}{*}{TCGA-RCC (\textcolor{blue}{R})}  & clear cell renal cell carcinoma (CC) & 498  \\
   & papillary renal cell (P)   & 289  \\
   & chromophobe renal cell carcinoma (ChRCC) & 118  \\
\midrule
\multirow{2}{*}{TCGA-NSCLC (\textcolor{blue}{N})}& squamous cell carcinoma (SCC) & 845  \\
   & adenocarcinoma (A) & 109  \\
\midrule
\multirow{2}{*}{TCGA-ESCA (\textcolor{red}{E})} & squamous cell carcinoma (SCC) & 114  \\
   & adenocarcinoma (A) & 86   \\
\midrule
\multirow{2}{*}{TCGA-TGCT (\textcolor{red}{T})} & seminoma (S) & 66   \\
   & mixed germ cell tumor (MGCT) & 29   \\
\midrule
\multirow{2}{*}{TCGA-CESC (\textcolor{red}{C})} & adenocarcinoma (A) & 270  \\
   & squamous cell carcinoma (SCC) &  49   \\
\bottomrule
\end{tabular}}
\caption{\textbf{Number of WSIs per cancer subtype for each TCGA cohort.} Abbreviations of rare cohorts are highlighted in \textcolor{red}{red}, while those of common cohorts are highlighted in {blue}.}
\label{tab:slide_counts}
\end{table}

\noindent \textbf{Metrics.} For evaluation, we use three primary accuracy metrics and two primary continual learning metrics.

\begin{table*}[]
\centering
\resizebox{1\linewidth}{!}{\begin{tabular}{l|c|lll|cc}
\toprule
\multicolumn{1}{c|}{\textbf{Method}} & \textbf{Buffer size} & \multicolumn{1}{c}{\textbf{\begin{tabular}[c]{@{}c@{}}{bACC} $\uparrow$\\ {(CLASS-IL)}\end{tabular}}} & \multicolumn{1}{c}{\textbf{\begin{tabular}[c]{@{}c@{}}{Masked bACC} $\uparrow$\\ {(TASK-IL)}\end{tabular}}} & \multicolumn{1}{c|}{\begin{tabular}[c]{@{}c@{}}\textbf{Mean ACC}$\uparrow$\\ \textbf{(CLASS-IL)}\end{tabular}} & \textbf{FGT}  $\downarrow$  & \textbf{\#BWT $\uparrow$}  \\ \midrule
Fully Supervised & \multirow{3}{*}{0 WSIs}  & {81.309 ($\pm$3.064)} & {89.688 ($\pm$1.687)} & \multicolumn{1}{c|}{--} & --  & -- \\
Naive Finetuning &  & {25.673 ($\pm$2.571)} & {86.597 ($\pm$1.933)} & 69.834 ($\pm$3.566)   & 4.649 ($\pm$2.558)  & -8.035 ($\pm$3.795) \\
Zero-shot & & {68.243 ($\pm$2.558)}$^{***}$ & {86.814 ($\pm$2.105)}$^{***}$ & 82.593 ($\pm$0.017)$^{***}$ & \textbf{0.909 ($\pm$0.406)} & -0.909 ($\pm$0.406) \\
\midrule
ER-ACE \cite{erace}  & \multirow{5}{*}{10 WSIs} & {44.354 ($\pm$4.280)} $^{***}$ & {67.407 ($\pm$5.282)} $^{***}$ & 71.663 ($\pm$3.265)$^{***}$ & \multicolumn{1}{l}{5.095 ($\pm$2.429)} & -0.320 ($\pm$1.211)  \\
AGEM \cite{agem} &  & \multicolumn{1}{l}{{42.076 ($\pm$4.287)}$^{***}$}& {80.364 ($\pm$2.206)}$^{***}$ & 74.514 ($\pm$4.016)$^{***}$ & \multicolumn{1}{l}{8.167 ($\pm$3.836)} & -7.714 ($\pm$3.978)   \\
DER++ \cite{derpp} &  & {52.033 ($\pm$4.272)} $^{***}$ & {84.735 ($\pm$2.407)}$^{***}$ & 81.786 ($\pm$1.575)$^{***}$ & \multicolumn{1}{l}{4.266 ($\pm$1.089)} & -3.627 ($\pm$1.182) \\
ConSlide \cite{huang2023conslide} &  & {62.731 ($\pm$4.488)}$^{***}$ & {87.027 ($\pm$2.553)}$^{***}$ & 81.992 ($\pm$1.055)$^{***}$ & \multicolumn{1}{l}{4.849 ($\pm$2.527)} &  \textbf{0.196 ($\pm$1.745)} \\ 
ADaFGrad \cite{adafgrad} &  & {65.267 ($\pm$2.235)}$^{***}$ & {89.458 ($\pm$2.299)}$^{***}$ & 83.651 ($\pm$1.755)$^{***}$ & \multicolumn{1}{l}{2.772 ($\pm$1.633)} &  -1.589 ($\pm$1.693) \\ \midrule
ER-ACE \cite{erace}  & \multirow{5}{*}{30 WSIs} & {58.556 ($\pm$5.654)}$^{***}$ & {75.255 ($\pm$5.161)}$^{***}$ & 81.138 ($\pm$2.464)$^{***}$ & \multicolumn{1}{l}{4.251 ($\pm$2.606)} &  \underline{0.088 ($\pm$3.317)} \\
AGEM \cite{agem} &  & {42.076 ($\pm$4.287)}$^{***}$ & {80.364 ($\pm$2.206)}$^{***}$ & 74.514 ($\pm$4.016)$^{***}$ & 8.167 ($\pm$3.836)  & -7.714 ($\pm$3.978) \\
DER++ \cite{derpp} &  & {55.560 ($\pm$2.746)}$^{***}$ & {86.863 ($\pm$2.810)}$^{***}$ & 83.580 ($\pm$1.532)$^{***}$ & 3.199 ($\pm$1.640) & -2.368 ($\pm$1.889) \\
ConSlide \cite{huang2023conslide}  &  & {64.622 ($\pm$1.755)}$^{***}$ & {88.346 ($\pm$1.977)}$^{***}$ & 82.602 ($\pm$1.201)$^{***}$ & \multicolumn{1}{l}{4.032 ($\pm$1.248)} & -2.930 ($\pm$1.506) \\
ADaFGrad \cite{adafgrad} &  & {69.034 ($\pm$3.861)}$^{***}$ & {\underline{89.704 ($\pm$2.082)}}$^{***}$ & 85.469 ($\pm$1.430)$^{***}$ & 2.517 ($\pm$1.125) & -1.217 ($\pm$1.805) \\ \midrule
\textbf{MergeSlide (ours) (naive)} & \multirow{2}{*}{0 WSIs} & {\underline{80.668 ($\pm$1.860)}} & {\textbf{92.087 ($\pm$1.740)}} & \underline{90.640 ($\pm$1.247)} & 4.941 ($\pm$1.518) & 5.384 ($\pm$1.317) \\ 
\textbf{MergeSlide (ours) \texttt{w/ TCP}} &  & {\textbf{87.929 ($\pm$2.110)}} & {\textbf{92.087 ($\pm$1.740)}} & \textbf{92.686 ($\pm$0.899)} & \underline{1.848 ($\pm$0.858)} & -1.604 ($\pm$1.024) \\
\bottomrule
\end{tabular}}
\caption{\textbf{Comparison between MergeSlide and other continual learning methods on a benchmark of six TCGA datasets in the sequence of \textcolor{blue}{B}$\rightarrow$\textcolor{blue}{R}$\rightarrow$\textcolor{blue}{N}$\rightarrow$\textcolor{red}{E}$\rightarrow$\textcolor{red}{T}$\rightarrow$\textcolor{red}{C}.} p-values between the best-performing method and the others: */**/*** indicate significance at the levels of $p < 0.05$, $p < 0.01$, and $p < 0.001$, respectively. {More results, including macro/weighted F1 and precision/recall/AUC per cancer subtype from this table, are reported in Table H.1 of the Supplementary Material}.}
\label{tab:main_results}
\end{table*}

\begin{table*}[]
\centering
\resizebox{1\linewidth}{!}{\begin{tabular}{l|c|lll|cc}
\toprule
\multicolumn{1}{c|}{\textbf{Method}} & \textbf{Buffer size} & \multicolumn{1}{c}{\textbf{\begin{tabular}[c]{@{}c@{}}{bACC} $\uparrow$\\ {(CLASS-IL)}\end{tabular}}} & \multicolumn{1}{c}{\textbf{\begin{tabular}[c]{@{}c@{}}{Masked bACC} $\uparrow$\\ {(TASK-IL)}\end{tabular}}} & \multicolumn{1}{c|}{\begin{tabular}[c]{@{}c@{}}\textbf{Mean ACC}$\uparrow$\\ \textbf{(CLASS-IL)}\end{tabular}} & \textbf{FGT}  $\downarrow$  & \textbf{\#BWT $\uparrow$}  \\ \midrule
Fully Supervised & \multirow{2}{*}{0 WSIs}  & {81.151 ($\pm$2.283)} & {90.138 ($\pm$1.854)} & 84.431 ($\pm$2.304) & --  & -- \\
Naive Finetuning &  & {24.976 ($\pm$3.330)} & {86.614 ($\pm$3.543)} & 42.565 ($\pm$1.031) & 7.561 ($\pm$4.773) & -6.982 ($\pm$4.557) \\ \midrule
DER++ \cite{derpp} & \multirow{3}{*}{30 WSIs} & {61.037 ($\pm$3.319)}$^{***}$ & {84.758 ($\pm$1.929)}$^{***}$ & 78.983 ($\pm$1.955)$^{***}$ & 6.644 ($\pm$4.062) & -5.676 ($\pm$4.490) \\
ConSlide \cite{huang2023conslide}  &  & {70.821 ($\pm$2.180)}$^{***}$ & {88.896 ($\pm$2.442)}$^{**}$ & 77.146 ($\pm$1.407)$^{***}$ & 3.117 ($\pm$1.784) & -2.521 ($\pm$1.940) \\
ADaFGrad \cite{adafgrad} &  & {77.096 ($\pm$4.474)}$^{***}$ & {\underline{90.216 ($\pm$1.677)}}$^{**}$ & 81.775 ($\pm$1.227)$^{***}$ & \underline{2.304 ($\pm$1.072)} & \underline{-0.821 ($\pm$0.918)} \\ \midrule
\textbf{MergeSlide (ours) (naive)} & \multirow{2}{*}{0 WSIs} & {\underline{80.636 ($\pm$1.865)}} & {\textbf{92.109 ($\pm$1.700)}} & \underline{88.268 ($\pm$1.459)} & 4.246 ($\pm$1.585) & -1.808 ($\pm$1.987) \\ 
\textbf{MergeSlide (ours) \texttt{w/ TCP}} &  & {\textbf{87.930 ($\pm$2.112)}} & {\textbf{92.109 ($\pm$1.700)}} & \textbf{91.009 ($\pm$1.278)} & \textbf{1.807 ($\pm$0.604)} & \textbf{0.958 ($\pm$0.868)} \\
\bottomrule
\end{tabular}}
\caption{\textbf{Comparison between MergeSlide and other continual learning methods on a benchmark of six TCGA datasets in the reversed sequence of \textcolor{red}{C}$\rightarrow$\textcolor{red}{T}$\rightarrow$\textcolor{red}{E}$\rightarrow$\textcolor{blue}{N}$\rightarrow$\textcolor{blue}{R}$\rightarrow$\textcolor{blue}{B}.} p-values between the best-performing method and the others: */**/*** indicate significance at the levels of $p < 0.05$, $p < 0.01$, and $p < 0.001$, respectively. {More results, including macro/weighted F1 and precision/recall/AUC per cancer subtype from this table, are reported in Table H.3 of the Supplementary Material.}}
\label{tab:reverse}
\end{table*}

\noindent\textbf{\textit{Accuracy metrics}} include: {1) Balanced Accuracy in the CLASS\mbox{-}IL setting (\textbf{bACC}), where the task identity is unknown to the model; 2) Balanced Accuracy in the TASK\mbox{-}IL setting (\textbf{Masked bACC}), where the model is given the task indicator $(X_i, t)$ and uses the correct $E^{\mathcal{C},t}$;} 3) \textbf{Mean ACC}: the mean of overall accuracy across a sequence of $T$ tasks. {Here, balanced accuracy is the average recall across classes}, and overall accuracy is the fraction of correct predictions over all samples in the test set. \noindent\textbf{\textit{Continual learning metrics}} include: 1) \textbf{Forgetting (FGT)}, which measures the decrease in CLASS\mbox{-}IL overall accuracy for task $t$ from its peak to its value after the final task; 2) \textbf{Backward Transfer (BWT)}, which quantifies how learning a new task affects performance on earlier tasks.

\noindent\textbf{{In/Out-of-domain Evaluation.}} {Because WSIs are acquired under different settings across institutes and sites, domain shift may occur \cite{cheng2021robust}, where a model overfitted to one site does not generalize well to another. To evaluate the stability of MergeSlide under domain shift, we design \textit{in-domain (IND)} and \textit{out-of-domain (OOD)} settings. In the \textit{in-domain} setting, training and testing slides may come from the same sites, while in the \textit{out-of-domain} setting, testing slides are drawn from different sites than those used for training. This simulates realistic scenarios where new slides from unseen sites must be tested on existing models.}

\noindent\textbf{Implemental Details.} WSIs are tiled into patches of size $256 \times 256$ at $10\times$ magnification. Patch embeddings ($v_j \in B'_i$) and prompt embeddings $e^{\mathcal{C},t}_j$ (from prompts $p^{\mathcal{C},t}_j \in P^{\mathcal{C},t}$) are extracted using TITAN's vision and text encoders, both producing 768-dimensional outputs across all methods. {Class-aware prompts used to extract embeddings are described in detail in Sec.~C of the Supplementary Material.} All experiments use the AdamW optimizer \cite{loshchilov2017decoupled} on a single RTX 8000 (48~GB) GPU. For ConSlide, we adopt the HIT backbone as in the original study, which supports both patch ($256\times256$) and region ($1024\times1024$) inputs. MergeSlide employs a Transformer aggregator $f_{\mathcal{A}}$, initialized with TITAN’s pretrained weights ($\theta_{{base}}$). All models are trained for $N_e = 10$ epochs. Rehearsal-based methods are evaluated with buffer sizes of 10 and 30 WSIs.

\noindent\textbf{Main Results (In-domain Results).} We choose the sequence \textcolor{blue}{B}$\rightarrow$\textcolor{blue}{R}$\rightarrow$\textcolor{blue}{N}$\rightarrow$\textcolor{red}{E}$\rightarrow$\textcolor{red}{T}$\rightarrow$\textcolor{red}{C} as the baseline. This sequence indicates that the model is finetuned from common to rare tasks. The main results are presented in Tab.~\ref{tab:main_results}. MergeSlide outperforms all rehearsal-based methods, achieving $\geq$18.899\% in CLASS-IL bACC, $\geq$7.217\% in Mean ACC, and $\geq$2.838\% in TASK-IL Masked bACC. Even without $\texttt{TCP}$ (considers all class-aware prompt embeddings), MergeSlide still outperforms other methods by $\geq$11.634\% and $\geq$5.171\% in terms of bACC and Mean ACC, respectively, under the CLASS-IL scenario. For forgetting (FGT), MergeSlide ranks second with –0.836\%, outperforming every trainable method (FGT $\geq$ 0.772\%) and trailing only the Zero-shot approach. Although its backward transfer (BWT) is slightly negative, it still delivers the best balance between stability and plasticity. 

\noindent\textbf{{Out-of-domain Results.}} {To assess domain shift, we evaluate all baselines and MergeSlide in the \textit{out-of-domain} setting; results are reported in Tab. \ref{tab:ood}. We additionally report the per-method difference between out-of-domain and in-domain balanced accuracy, denoted $\Delta^{\text{bACC}}_{out\text{-}in}$. Although MergeSlide shows a $-2.817\%$ bACC drop under domain shift (as do AGEM, ConSlide, and ADaFGrad), it still secures the best bACC among all methods in both CLASS-IL ($\geq$23.276\% bACC) and TASK-IL ($\geq$3.465\% bACC) settings, and it also leads in FGT ($\geq$1.304\%).}

\begin{table}[]
\resizebox{1\linewidth}{!}{{\begin{tabular}{lllcc}
\toprule
{\textbf{Method}}     & \textbf{\textbf{\begin{tabular}[c]{@{}c@{}}{bACC} $\uparrow$\\ {(CLASS-IL)}\end{tabular}}}           & \textbf{\begin{tabular}[c]{@{}c@{}}{Masked bACC} $\uparrow$\\ {(TASK-IL)}\end{tabular}}    & \textbf{{FGT}$\downarrow$}            & \textbf{{$\Delta^{\text{bACC}}_{out-in}$}$\downarrow$}             \\ \midrule
{ER-ACE} \cite{erace} & {60.808 ($\pm$4.769)}$^{***}$ & {77.125 ($\pm$4.800)}$^{***}$ & {4.447 ($\pm$2.431)}  & {+2.252}  \\
{AGEM} \cite{agem}   & {41.864 ($\pm$4.186)}$^{***}$ & {76.207 ($\pm$3.793)}$^{***}$ & {12.957 ($\pm$4.964)} & {-0.212} \\
{DER++} \cite{derpp}      & {57.951 ($\pm$3.115)}$^{***}$ & {83.469 ($\pm$2.902)}$^{***}$ & {3.994 ($\pm$1.584)}  & {+2.391}  \\
{ConSlide} \cite{huang2023conslide}   & {61.336 ($\pm$5.791)}$^{***}$ & {85.945 ($\pm$4.393)}$^{***}$ & {3.214 ($\pm$1.967)}  & {-3.286}  \\
{ADaFGrad} \cite{adafgrad} & {61.836 ($\pm$5.562)}$^{***}$ & {86.110 ($\pm$3.217)}$^{***}$ & {4.108 ($\pm$2.603)}  & {-7.198}  \\ \midrule\midrule
{MergeSlide (ours)} & \textbf{{85.112 ($\pm$1.570)}} & \textbf{{89.575 ($\pm$2.440)}} & \textbf{{1.910 ($\pm$0.089)}} & \textbf{{-2.817}} \\ \bottomrule
\end{tabular}}}
\caption{{\textbf{Comparison between MergeSlide and other continual learning methods on \textbf{out-of-domain setting} in the sequence of \textcolor{blue}{B}$\rightarrow$\textcolor{blue}{R}$\rightarrow$\textcolor{blue}{N}$\rightarrow$\textcolor{red}{E}$\rightarrow$\textcolor{red}{T}$\rightarrow$\textcolor{red}{C}.} {More results are reported in Table H.2.}} \vspace{-0.7cm}}
\label{tab:ood}
\end{table}

\section{Additional Analyses}


\noindent\textbf{Varying Task Orders.} First, we evaluate the rare-to-common sequence, i.e., \textcolor{red}{C}$\rightarrow$\textcolor{red}{T}$\rightarrow$\textcolor{red}{E}$\rightarrow$\textcolor{blue}{N}$\rightarrow$\textcolor{blue}{R}$\rightarrow$\textcolor{blue}{B}. High-performing methods from the main results are tested under the reversed task order, with results in Tab.~\ref{tab:reverse} showing that rehearsal-based methods are sensitive to task order. For example, DER++, ConSlide, and ADaFGrad perform better when rare tasks are presented before common ones. In contrast, \textit{MergeSlide maintains stable performance with TCP and shows only marginal degradation in bACC (-0.032\%) and Mean ACC (-1.677\%) without TCP}. We also examine alternating rare and common tasks (e.g., \textcolor{blue}{B}$\rightarrow$\textcolor{red}{C}$\rightarrow$\textcolor{blue}{R}$\rightarrow$\textcolor{red}{T}$\rightarrow$\textcolor{blue}{N}$\rightarrow$\textcolor{red}{E}), with four variations tested. As shown in Tab.~\ref{tab:sequence}, \textit{MergeSlide consistently maintains stable performance across sequences.}

\begin{table}[]
\resizebox{\linewidth}{!}{\begin{tabular}{cccc}
\toprule
\textbf{Sequence} & \textbf{{ACC}} & \textbf{{Masked ACC}} & \textbf{Mean ACC} \\ \midrule
\textcolor{blue}{B}$\rightarrow$\textcolor{red}{C}$\rightarrow$\textcolor{blue}{R}$\rightarrow$\textcolor{red}{T}$\rightarrow$\textcolor{blue}{N}$\rightarrow$\textcolor{red}{E} & 91.955 ($\pm$0.902) &  93.637 ($\pm$1.084) & 91.861 ($\pm$0.914) \\
\textcolor{red}{E}$\rightarrow$\textcolor{blue}{N}$\rightarrow$\textcolor{red}{T}$\rightarrow$\textcolor{blue}{R}$\rightarrow$\textcolor{red}{C}$\rightarrow$\textcolor{blue}{B} & 91.975 ($\pm$0.915) & 92.433 ($\pm$1.084) & 91.381 ($\pm$1.126) \\
\textcolor{red}{C}$\rightarrow$\textcolor{blue}{B}$\rightarrow$\textcolor{red}{T}$\rightarrow$\textcolor{blue}{R}$\rightarrow$\textcolor{red}{E}$\rightarrow$\textcolor{blue}{N} &  91.933 ($\pm$1.061) & 93.637 ($\pm$1.084) & 91.336 ($\pm$1.154) \\
\textcolor{blue}{N}$\rightarrow$\textcolor{red}{E}$\rightarrow$\textcolor{blue}{R}$\rightarrow$\textcolor{red}{T}$\rightarrow$\textcolor{blue}{B}$\rightarrow$\textcolor{red}{C} & 91.955 ($\pm$0.926) & 93.616 ($\pm$1.098) & 91.021 ($\pm$1.041)  \\ \midrule
$\sigma$ & 0.0149 & 0.5184 & 0.3003 \\
\bottomrule
\end{tabular}}
\caption{\textbf{Experiments on alternative task orders where rare and common tasks are placed alternately.} Column $\sigma$ indicates the standard deviation of average results across three metrics.}
\label{tab:sequence}
\end{table}

\begin{figure}[!http]
\centerline{\includegraphics[width=0.49\textwidth]{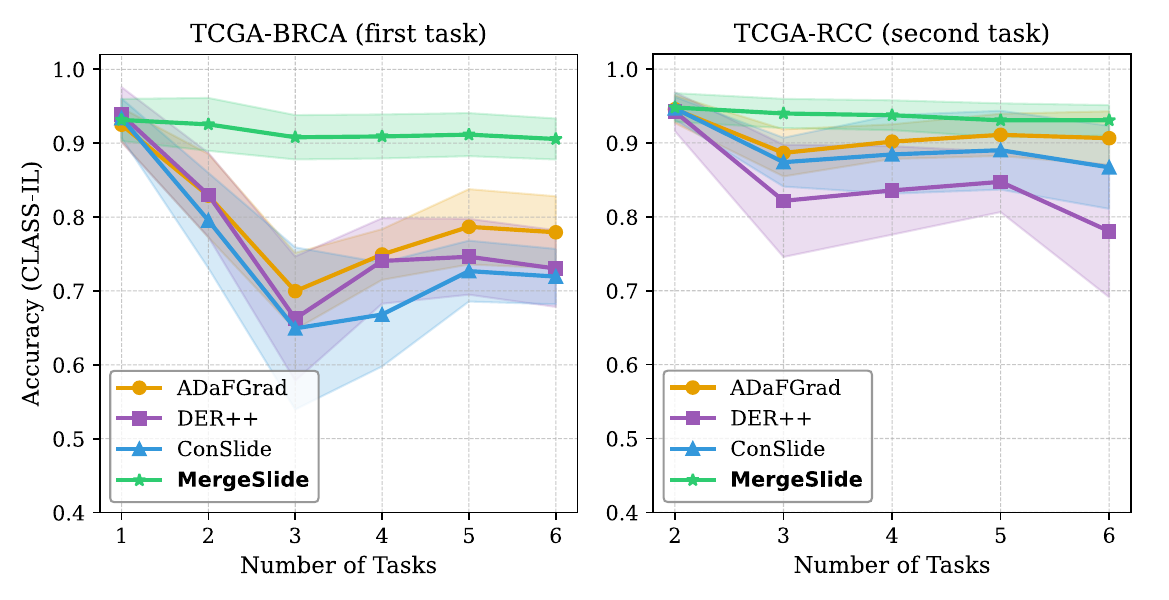}}
\caption{\textbf{Performance drop comparisons across different methods when new tasks are added on TCGA-BRCA and TCGA-RCC.} \vspace{-0.5cm}}
\label{fig:forgetting} 
\end{figure}

\begin{figure}[!http]
\centerline{\includegraphics[width=0.49\textwidth]{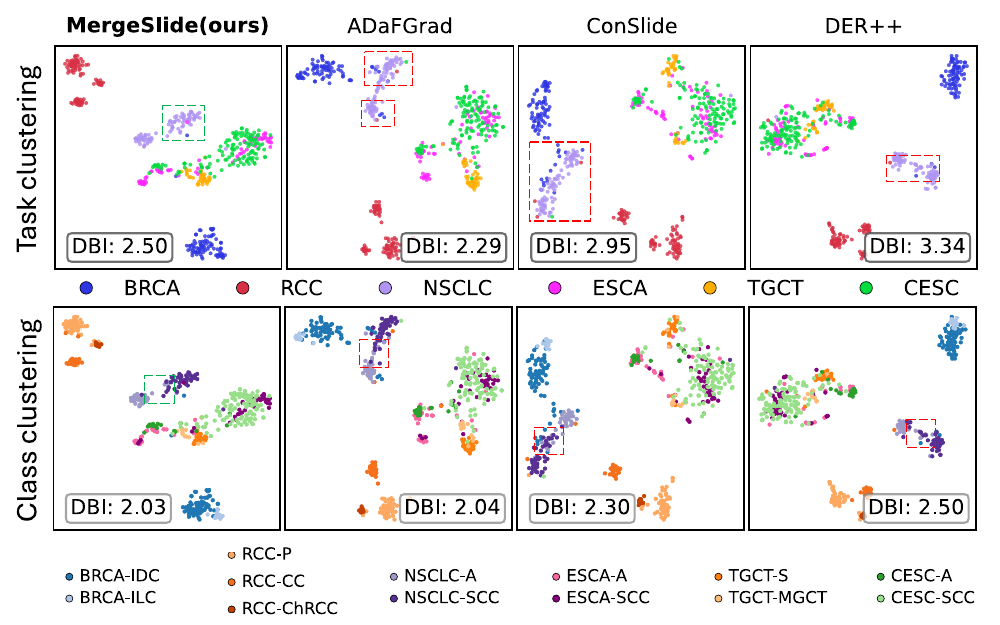}}
\caption{\textbf{t-SNE \cite{van2008visualizing} visualization of slide embeddings from MergeSlide and three comparative methods, organized by task and class spaces.} The dashed \textcolor{teal}{green} box highlights MergeSlide’s superior clustering, compared to the regions marked by dashed \textcolor{red}{red} boxes in other methods.}
\label{fig:tsne} 
\end{figure}

\noindent\textbf{Qualitative Analyses.} To take a closer look at how well tasks and cancer subtypes are recognized after training on the last task, we provide a t-SNE visualization in Fig.~\ref{fig:tsne} to show how the embeddings are clustered at both the task and class levels across the top selected methods. To further highlight the differences, we also compute the Davies–Bouldin Index (DBI) \cite{davies2009cluster}, where a lower value indicates better clustering performance. In task-level clustering, MergeSlide demonstrates superior performance on the NSCLC dataset, which is well recognized, with only two samples from ESCA and BRCA misclustered. In contrast, other methods show significant misclustering, with WSIs from the NSCLC dataset frequently mixed with those from BRCA, CESC, and RCC. In class-level clustering, IDC and ILC cancer subtypes are better separated by MergeSlide, while other methods tend to mix them together.

\noindent{{\textbf{Complexity Analysis of MergeSlide.} Since the slide aggregator $f_{\mathcal{A}}$ includes $L$ layers, each with weights $\theta_l \in \mathbb{R}^{m_l \times n_l}$, the overall time complexity of merging $|\mathcal{D}|$ tasks is $\mathcal{O}(|\mathcal{D}|\sum_{l=1}^{L} m_l n_l^2)$. Details are provided in Sec. I.1 of the supplementary. For empirical evaluation, Tab.~\ref{tab:complexity} reports the computational profile of MergeSlide and two strongest baselines on a single NVIDIA RTX 4090 GPU, using three metrics: average epoch time per task, GPU VRAM usage, and inference throughput. The measurements exclude WSI preprocessing (segmentation, feature extraction), as it is identical across methods. During training, MergeSlide requires more VRAM due to the orthogonal merging strategy, but its epoch time matches ConSlide and is $1.3\times$ faster than ADaFGrad. At inference, MergeSlide shows no difference with or without TCP, incurs only a slight overhead compared to ConSlide/ADaFGrad ($+0.28$s), and achieves a throughput lower by 12.07 slides/s. This trade-off remains acceptable given the substantial performance gains.}


\begin{table}[]
\resizebox{1\linewidth}{!}{{
\begin{tabular}{cllccc}
\toprule
\textbf{Stage}             & \multicolumn{2}{c}{\textbf{Method}}               & \textbf{Avg. Time (s)} & \textbf{VRAM (GB)} & \textbf{Slides/s} \\ \midrule
\multirow{5}{*}{\rotatebox[origin=c]{90}{Training}}  
  & \multirow{3}{*}{MergeSlide} & Per-task FT & 74.26 & 3.38 & --  \\
  & & Merging   & 4.50 & 0.71 & --  \\ \cmidrule{3-6}
  & & Total     & 78.76 & 4.09 & --  \\ \cmidrule{2-6}
  & \multicolumn{2}{l}{ConSlide \cite{huang2023conslide}} & 78.15 & 2.90 & --  \\
  & \multicolumn{2}{l}{ADaFGrad \cite{adafgrad}}          & 104.17 & 2.70 & --  \\ \midrule
\multirow{3}{*}{\rotatebox[origin=c]{90}{Inference}} 
  & \multicolumn{2}{l}{MergeSlide (naive)} & 1.52 & 3.80 & 55.81 \\
  & \multicolumn{2}{l}{MergeSlide \texttt{\textbf{w/ TCP}}} & 1.52 & 3.89 & 55.82 \\ \cmidrule{2-6}
  & \multicolumn{2}{l}{ConSlide \cite{huang2023conslide}/ADaFGrad \cite{adafgrad}} & 1.24 & 0.87 & 67.89 \\ \bottomrule
\end{tabular}
}}
\caption{{\textbf{Computational comparison of average epoch time (s), GPU VRAM (GB), and throughput (slides/s).} MergeSlide training includes Per-task Fine-tuning (FT) and Orthogonal Model Merging. ConSlide/ADaFGrad follow the same flow during inference.}\vspace{-0.5cm}}
\label{tab:complexity}
\end{table}

\section{Conclusion}
\label{sec:conclusion}
MergeSlide enables simple yet effective lifelong learning on WSIs through class-aware prompt design, MLP-free backbone fine-tuning, and model merging. Its task-level, prompt-aligned inference leads to more accurate predictions under the CLASS-IL setting. The superior performance over both vision-language zero-shot and rehearsal-based methods demonstrates its effectiveness. Additionally, MergeSlide remains stable across different task order permutations as the number of tasks increases. 

\noindent{\textbf{Limitations \& Perspectives.} MergeSlide relies on class-aware prompts and pathology VLMs, whose textual descriptions should be carefully designed. For unseen generalization, it works naturally when new tasks arise, but handling unseen classes within existing tasks requires strategies such as class updates or out-of-distribution detection. In addition, the current orthogonal model merging uses full SVD, which may be time-consuming as tasks scale; exploring low-rank SVD is a promising direction. The performance of MergeSlide on broader tasks (e.g., cancer grades, survival bins) will also be investigated.}

{\small
\bibliographystyle{ieee_fullname}
\bibliography{main}
}



\section*{Supplementary material}

# MergeSlide: Continual Model Merging and Task-to-Class Prompt-Aligned Inference for Lifelong Learning on Whole Slide Images
*Supplementary Material*

Doanh C. Bui[1,*], Ba Hung Ngo[2], Hoai Luan Pham[1], Khang Nguyen[3], Maï K. Nguyen[4], Yasuhiko Nakashima[1]

[1]Nara Institute of Science and Technology, Japan
[2]Graduate School of Data Science, Chonnam National University, South Korea
[3]University of Information Technology, Viet Nam National University Ho Chi Minh City
[4]ETIS (UMR 8051), CY Cergy Paris University, ENSEA, CNRS, France

bui.cao_doanh.bd2@naist.ac.jp

## Contents

## A. Overview

In the supplementary material, we provide pseudocode for MergeSlide (Sec. B), implementation details of the backbone $f_\mathcal{A}$ (Sec. C), the full set of designed class-aware prompts per task (Sec. D), clarification of the class-incremental and task-incremental scenarios (Sec. E), definitions of evaluation metrics (Sec. F), and a discussion on comparison with the zero-shot approach (Sec. G). We further report additional metrics, including per-class Precision, Recall, AUC, and Macro/Weighted F1 (Sec. H). Additional comparisons are also included (Sec. I), such as task-wise performance drops, confidence score analysis, an ablation study on task-specific MLP-free fine-tuning, and performance after training on the final task.

## B. Pseudocode

To provide a clearer description of MergeSlide, we present Algorithm 1, which comprehensively summarizes its processes and operations when training on a stream of $T$ datasets.

## C. Details of Slide Aggregators

In this section, we provide details of the slide aggregator $f_\mathcal{A}$ used in this study. For other rehearsal-based methods such as ER-ACE [3], AGEM [4], DER++ [2], ADaFGrad [1], and ConSlide [7], $f_\mathcal{A}$ is implemented using the HIT backbone introduced in the ConSlide study [7] to ensure a fair comparison. HIT is a hierarchical transformer designed to leverage pyramid multi-level images of a WSI. This network includes $L = 2$ Transformer blocks with a hidden dimension size of $d_{\text{model}} = 384$. Each Transformer block contains two self-attention layers for patch-level and region-level embeddings, respectively, with a cross-interaction mechanism to

**Algorithm 1** MergeSlide

1: **Input:** Sequence of $T$ training datasets $\mathcal{D}^{train} = \{D_t^{train}\}_{t=1}^T$, each with $c_t$ cancer subtypes. Testing datasets: $\mathcal{D}^{test} = \{D_t^{test}\}_{t=1}^T$.
2: **Init:** Pre-trained slide aggregator $f_\mathcal{A}(\cdot, \theta_{base})$
3: Initialize controlling factor $\lambda_1 \leftarrow 1$
4: $\forall X_i \in \{D_t^{train}, D_t^{test}\}_{t=1}^T$, perform tissue segmentation, tiling, patch embedding, and sampling to obtain $B_i'$.
5: **Fine-tuning and model merging:**
6: **for** $t = 1$ to $T$ **do**
7:     Fine-tuning for task $t$:    ▷ Sec. 3.2.1
8:     Define class prompts $P^{\mathcal{C},t}$ and embeddings $E^{\mathcal{C},t}$
9:     Initialize weights: $\theta_t \leftarrow \theta_{base}$
10:    Train $f_\mathcal{A}$ on $B_i' \in D_t^{train}$:
11:       $Z_i \leftarrow f_\mathcal{A}(B_i', \theta_t)$
12:       $\hat{p}_i \leftarrow \{Z_i \cdot (e_j^{\mathcal{C},t})^\top \mid e_j^{\mathcal{C},t} \in E^{\mathcal{C},t}\}$
13:    Update: $\theta_t \leftarrow \arg\min_{\theta_t} \mathcal{L}_t(y_i, \hat{p}_i)$
14:    Continual model merging:    ▷ Sec. 3.2.2
15:    **if** $t > 1$ **then**
16:       $\Delta\tilde{\theta}_{1:t-1} \leftarrow \tilde{\theta}_{1:t-1} - \theta_{base}$
17:       $\Delta\theta_t \leftarrow \theta_t - \theta_{base}$
18:       SVD: $\Delta\tilde{\theta}_{1:t-1} \leftarrow U\Sigma V^\top$
19:       Project: $G(\Delta\theta_t) \leftarrow U\left((U^\top \Delta\theta_t V) \odot M\right) V^\top$    ▷ Eq. (4)
20:       Compute $\lambda_t \leftarrow t \cdot \frac{\|\lambda_{t-1}\Delta\tilde{\theta}_{1:t-1}+G(\Delta\theta_t)\|_2}{\sum_{i=1}^t \|\theta_i\|_2}$    ▷ Eq. (5)
21:       Merge: $\tilde{\theta}_{1:t} \leftarrow \theta_{base} + \frac{\lambda_{t-1}\Delta\tilde{\theta}_{1:t-1}+G(\Delta\theta_t)}{\lambda_t}$    ▷ Eq. (6)
22:    **else**
23:       $\tilde{\theta}_{1:t} \leftarrow \theta_t$
24:    **end if**
25: **end for**
26: **Inference on all test sets:**
27: Compute task-level prompts: $E^{\mathcal{T},t} \leftarrow \frac{1}{c_t} \sum E^{\mathcal{C},t}$ for all $t$
28: $\mathcal{E}^\mathcal{T} \leftarrow \{E^{\mathcal{T},t}\}_{t=1}^T$
29: **for** $t = 1$ to $T$ **do**
30:    **for** $B_i'$ in $D_t^{test}$ **do**
31:       $Z_i \leftarrow f_\mathcal{A}(B_i', \tilde{\theta}_{1:T})$
32:       CLASS-IL (naive):
33:       $\hat{p}_i \leftarrow \{\}$
34:       **for** $t = 1$ to $T$ **do**
35:          $\hat{p}_i \leftarrow \hat{p}_i \cup \{Z_i \cdot (e_j^{\mathcal{C},t})^\top \mid e_j^{\mathcal{C},t} \in E^{\mathcal{C},t}\}$
36:       **end for**
37:       Predict: $\hat{y}_i \leftarrow \arg\max \hat{p}_i$
38:       CLASS-IL with **prompt-aligned inference**:    ▷ Sec. 3.2.3
39:       $\hat{t} \leftarrow \arg\max_t\{Z_i \cdot (E^{\mathcal{T},t})^\top\}$
40:       Predict: $\hat{y}_i \leftarrow \arg\max_j\{Z_i \cdot (e_j^{\mathcal{C},\hat{t}})^\top\}$
41:       Task-IL:
42:       Predict: $\hat{y}_i \leftarrow \arg\max_j\{Z_i \cdot (e_j^{\mathcal{C},t})^\top\}$
43:    **end for**
44: **end for**
45: **Return:** Final unified model weights $\tilde{\theta}_{1:T}$

fuse these two levels. Specifically, region embeddings are added to each patch embedding. All patch embeddings are then passed through a convolutional layer and max-pooled before being added back to the region embedding. A region

comprises multiple corresponding patches. In this study, we use region images (of size $1024 \times 1024$) at $2.5\times$ magnification and patch images (of size $256 \times 256$) at $10\times$, meaning each region contains $4 \times 4$ patch images. For the Zero-shot setting and MergeSlide, since $f_\mathcal{A}$ should be pre-trained to contain strong base knowledge $\theta_{\text{base}}$, we adopt the pre-trained Transformer-based network introduced in TITAN [6], which is a variant of the Vision Transformer that with $d_{\text{model}} = 768$. However, in MergeSlide, we sample only $K = 400$ patches per slide, whereas other methods utilize all patches to achieve the best performance.

## D. How Class-aware Prompts Are Designed?

The class-aware prompts are designed based on the following format: class template + [CLASSNAME], where the [CLASSNAME] tag is a short natural-language description of a cancer subtype. We leveraged 22 class template variants, as shown in Table D.1. For each cancer subtype, there are 3~4 class tags, listed in Table D.2. For each $i$-th cancer subtype, variants of class-aware prompts is constructed by concatenating all 22 class templates with the 3~4 class tags, resulting in approximately 88 class-aware prompts (denoted as $p_i^{\mathcal{C},t}$) for each cancer subtype. These prompts are then passed through the text encoder of TITAN to obtain 88 embedding vectors, which are averaged across dimensions to produce a class-aware embedding vector $e_i^{\mathcal{C},t} \in \mathbb{R}^{d_{text}}$. For each $t$-th dataset/task, there are $c_t$ cancer subtypes (e.g., for TCGA-BRCA, there are two cancer subtypes: invasive ductal carcinoma and invasive lobular carcinoma). A set of class-aware prompts is then constructed as $E^{\mathcal{C},t} = \{e_i^{\mathcal{C},t}\}_{i=1}^{c_t}$, which is used to retrieve the final prediction by computing the dot-product similarity with the slide embedding.

| # | Prompt template |
|---|---|
| ❶ | "[CLASSNAME]." |
| ❷ | "a photomicrograph showing [CLASSNAME]." |
| ❸ | "a photomicrograph of [CLASSNAME]." |
| ❹ | "an image of [CLASSNAME]." |
| ❺ | "an image showing [CLASSNAME]." |
| ❻ | "an example of [CLASSNAME]." |
| ❼ | "[CLASSNAME] is shown." |
| ❽ | "this is [CLASSNAME]." |
| ❾ | "there is [CLASSNAME]." |
| ❿ | "a histopathological image showing [CLASSNAME]." |
| ⓫ | "a histopathological image of [CLASSNAME]." |
| ⓬ | "a histopathological photograph of [CLASSNAME]." |
| ⓭ | "a histopathological photograph showing [CLASSNAME]." |
| ⓮ | "shows [CLASSNAME]." |
| ⓯ | "presence of [CLASSNAME]." |
| ⓰ | "[CLASSNAME] is present." |
| ⓱ | "an H&E stained image of [CLASSNAME]." |
| ⓲ | "an H&E stained image showing [CLASSNAME]." |
| ⓳ | "an H&E image showing [CLASSNAME]." |
| ⓴ | "an H&E image of [CLASSNAME]." |
| ㉑ | "[CLASSNAME], H&E stain." |
| ㉒ | "[CLASSNAME], H&E." |

Table D.1. Prompt template variations.

| # | Cancer Subtype | [CLASSNAME] |
|---|---|---|
| TCGA-BRCA | invasive ductal carcinoma | ❶ "invasive ductal carcinoma" <br> ❷ "breast invasive ductal carcinoma" <br> ❸ "invasive ductal carcinoma of the breast" <br> ❹ "invasive carcinoma of the breast, ductal pattern" |
| TCGA-BRCA | invasive lobular carcinoma | ❶ "invasive lobular carcinoma" <br> ❷ "breast invasive lobular carcinoma" <br> ❸ "invasive lobular carcinoma of the breast" <br> ❹ "invasive carcinoma of the breast, lobular pattern" |
| TCGA-RCC | clear cell | ❶ "clear cell renal cell carcinoma" <br> ❷ "renal cell carcinoma, clear cell type" <br> ❸ "renal cell carcinoma of the clear cell type" <br> ❹ "clear cell rcc" |
| TCGA-RCC | papillary | ❶ "papillary renal cell carcinoma" <br> ❷ "renal cell carcinoma, papillary type" <br> ❸ "renal cell carcinoma of the papillary type" <br> ❹ "papillary rcc" |
| TCGA-RCC | chromophobe renal cell carcinoma | ❶ "chromophobe renal cell carcinoma" <br> ❷ "renal cell carcinoma, chromophobe type" <br> ❸ "renal cell carcinoma of the chromophobe type" <br> ❹ "chromophobe rcc" |
| TCGA-NSCLC | squamous cell carcinoma | ❶ "squamous cell carcinoma" <br> ❷ "lung squamous cell carcinoma" <br> ❸ "squamous cell carcinoma of the lung" <br> ❹ "lusc" |
| TCGA-NSCLC | adenocarcinoma | ❶ "adenocarcinoma" <br> ❷ "lung adenocarcinoma" <br> ❸ "adenocarcinoma of the lung" <br> ❹ "luad" |
| TCGA-ESCA | squamous cell carcinoma | ❶ "squamous cell carcinoma" <br> ❷ "esophageal squamous cell carcinoma" <br> ❸ "squamous cell carcinoma of the esophagus" <br> ❹ "escc" |
| TCGA-ESCA | adenocarcinoma | ❶ "adenocarcinoma" <br> ❷ "esophageal adenocarcinoma" <br> ❸ "adenocarcinoma of the esophagus" <br> ❹ "esad" |
| TCGA-TGCT | Sinoma | ❶ "Sinoma" <br> ❷ "testicular Sinoma" <br> ❸ "Sinoma of the testis" |
| TCGA-TGCT | mixed germ cell tumor | ❶ "mixed germ cell tumor" <br> ❷ "testicular mixed germ cell tumor" <br> ❸ "mixed germ cell tumor of the testis" |
| TCGA-CESC | squamous cell carcinoma | ❶ "squamous cell carcinoma" <br> ❷ "cervical squamous cell carcinoma" <br> ❸ "squamous cell carcinoma of the cervix uteri" |
| TCGA-CESC | adenocarcinoma | ❶ "adenocarcinoma" <br> ❷ "cervical adenocarcinoma" <br> ❸ "adenocarcinoma of the cervix uteri" |

Table D.2. Class-aware prompts for each cancer subtype.

## E. Explanations of CLASS-IL and TASK-IL

Under the CLASS-IL scenario, the model does not know the task identity (i.e., $t$ is unknown) of an input WSI, whereas under the TASK-IL scenario, the model knows which task should be targeted (i.e., $t$ is known).

### E.1. Class-imcremental Learning (CLASS-IL)

Following the iCARL implementation [8], all continual learning methods take the argmax of the predicted logits $\hat{y} = \arg\max \hat{p}_i$, where $\hat{p} \in \mathbb{R}^C$ and $C$ is the total number of classes seen so far from $T$ tasks, i.e., $C = \sum_{t=1}^{T} c_t$, as the final prediction. For MergeSlide, in the naive setting, we compute all dot-product similarities between the input slide embedding and all class-aware prompt embeddings $\{E^{\mathcal{C},t}\}_{t=1}^{T}$ to obtain $\hat{p}_i$ (as shown in lines 31–34 of Algorithm 1). Under TCP inference, the model is first given the task identity $\hat{t}$ by selecting the task-level prompt embedding $\{E^{\mathcal{T},t}\}_{t=1}^{T}$ that yields the highest similarity. After obtain-

ing $\hat{t}$, the corresponding set of class-aware prompt embeddings $E^{\mathcal{C},\hat{t}}$ is used to compute the final prediction. *We note that this strategy is still valid under the CLASS-IL setting, where the model does not know the task identity and must consider all cancer subtype classes seen so far. The TCP inference strategy identifies the task using a vision-language similarity approach, thereby narrowing the task boundary. However, it does not assume access to the task identity at the beginning.*

### E.2. Task-imcremental Learning (TASK-IL)

Under the TASK-IL scenario, the model is provided with the task identity, and therefore knows the task boundary. Specifically, given task $t$, the task boundary is defined as $s_t = \sum_{i=1}^{t-1} c_i$, $e_t = \sum_{i=1}^{t} c_i$, where $c_i$ is the number of classes of $i$-th task. Then, the prediction is computed by taking $\hat{y} = \arg\max_{j \in s_t, \ldots, e_t} \hat{p}_{i,j}$, where $\hat{p}_{i,j}$ is the $j$-th logit representing the $j$-th cancer subtype class after training on all tasks. For MergeSlide, as the model knows the task identity, we simply pick the corresponding set of class-aware prompts $E^{\mathcal{C},t}$ to retrieve the final prediction.

## F. Explanations of Metrics

Alongside (1) **bACC** and (2) **Masked bACC**, we evaluate continual learning performance using three metrics: (3) **Mean Accuracy (mACC)**, (4) **Backward Transfer (BWT)**, (5) **Forgetting (FGT)**.

### F.1. Balanced Accuracy (bACC)

Since each dataset in the six-TCGA benchmark is highly imbalanced, we report Balanced Accuracy as the main metric for both CLASS-IL and TASK-IL scenarios to better reflect performance on such test sets. bACC of the $t$-th task, which involves classifying $c_t$ classes, is defined as:

$$bACC_t = \frac{1}{c_t} \sum_{i=1}^{c_t} \frac{TP_i}{TP_i + FN_i}, \quad (1)$$

where $TP_i$ is the number of samples correctly classified as the $i$-th cancer subtype in the $t$-th task, and $FN_i$ is the number of samples of the $i$-th cancer subtype misclassified as other subtypes. In other words, this corresponds to the macro-averaged recall across all cancer subtypes. The final bACC is computed as the mean bACC across all $T$ tasks:

$$bACC = \frac{1}{T} \sum_{t=1}^{T} bACC_t. \quad (2)$$

### F.2. Accuracy (ACC)

Accuracy is defined as the ratio of correctly classified samples to the total number of samples in a task, and it has been the primary evaluation metric in previous continual learning studies [1–4, 7]. To ensure fair comparison with these works, our study also adopts accuracy as the base metric for computing other continual learning measures.

Let $T$ be the total number of tasks, and let $ACC_i(j)$ denote the accuracy on task $j$ after training up to task $i$. The equations for **Mean ACC**, **BWT**, and **FGT** are provided in Secs. F.3, F.4, and F.5, respectively.

### F.3. Mean Accuracy (Mean ACC)

Mean Accuracy is defined as the average of CLASS-IL accuracies over $T$ sequences of incrementally learned tasks:

$$\text{mACC} = \frac{1}{T} \sum_{t=1}^{T} ACC_t, \quad (3)$$

where $ACC_t$ denotes the accuracy under the CLASS-IL scenario on all tasks up to and including the $t$-th task.

### F.4. Backward Transfer (BWT)

Backward Transfer measures how learning new tasks affects performance on old tasks:

$$\text{BWT} = \frac{1}{T-1} \sum_{t=1}^{T-1} \left( ACC_T(t) - ACC_t(t) \right). \quad (4)$$

### F.5. Forgetting (FGT)

Forgetting measures the drop from each task's best achieved accuracy to its final accuracy after all training:

$$\text{FGT} = \frac{1}{T-1} \sum_{t=1}^{T-1} \left( \max_{i \in \{t, \ldots, T\}} ACC_i(t) - ACC_T(t) \right). \quad (5)$$

For ACC, Masked ACC, mACC, BWT, higher values indicate better performance, while for FGT, lower values are better. Specifically, BWT measures how tasks influence each other during the training process. Therefore, we expect the model to facilitate positive transfer between tasks; in other words, higher values are desirable. In contrast, FGT measures how much the model forgets old tasks after training on new ones, and this should be minimized. That is, lower values are better.

### F.6. Precision, Recall, AUC and F1

To further investigate per-class performance across all models, we compute Precision, Recall, and AUC for each cancer subtype, as well as Weighted and Macro F1. For Precision, Recall, and AUC, we aggregate the predictions of all WSIs across $T$ tasks and then compute the metrics to evaluate how well each cancer subtype can be recognized among all possible subtypes. Weighted and Macro F1 are also computed across all cancer subtypes in the sequence of $T$ testing sets. The results are reported in Sec. H.

### F.7. Davies-Bouldin Index (DBI)

In Qualitative Analyses provided in the main manuscript (Sec. 5), we also compute DBI to measure how good each slide embeddings clustered to its task/class spaces.

$$\text{DBI} = \frac{1}{K} \sum_{i=1}^{K} \max_{j \neq i} \left( \frac{S_i + S_j}{M_{ij}} \right), \quad (6)$$

where $S_i = \frac{1}{|C_i|} \sum_{Z \in C_i} \|Z - \mu_i\|^2$, where $Z$ is the slide embedding of a WSI aggregated by $f_\mathcal{A}$, and $C_i$ is the $i$-th cluster, i.e., the set of WSIs belonging to the same task or cancer subtyping class. $\mu_i$ is the centroid of cluster $i$, and $M_{ij} = \|\mu_i - \mu_j\|^2$ denotes the squared distance between the $i$-th and $j$-th centroids. Although t-SNE shows points in a projected 2D space, DBI is still computed on the original $d$-dimensional slide embedding.

## G. Discussion For Zero-shot Approach

For zero-shot evaluation, we simply obtain the embedding from the pre-trained $f_\mathcal{A}$ without any task-specific training, and compute the dot-product similarity with all class-aware prompt embeddings in $\mathcal{E}^\mathcal{C}$. The resulting scores are then normalized using the softmax function, and the class with the highest score is selected as the final prediction. Let $Z_i$ be the slide embedding produced by $f_\mathcal{A}$, and let the initial prediction vector be $\hat{p}_i = \{\varnothing\}$. This process can be expressed as:

$$\begin{aligned} \hat{p}_i &= \hat{p}_i \cup \{Z_i \cdot (e_j^{\mathcal{C},t})^\top \mid \forall e_j^{\mathcal{C},t} \in E^{\mathcal{C},t}\}, \quad \text{for } t = 1, \dots, T \\ \hat{y}_i &= \arg\max (\text{softmax}(\hat{p}_i)) \end{aligned} \quad (7)$$

We note that a head-to-head comparison with the zero-shot setting is important, as MergeSlide relies on pathology VLMs. One might expect that zero-shot inference could achieve good performance without any task-specific training, leveraging vision-language alignment conditioned on well-crafted class-aware prompts. However, we show that simply relying on this approach does not outperform rehearsal-based continual learning methods, although it remains competitive. In contrast, our study proposes training an MLP-free slide aggregator $f_\mathcal{A}$, performing prediction using class-aware prompt embeddings over a few epochs, and then merging the results into a unified model. This approach significantly outperforms the zero-shot setting while maintaining a favorable trade-off between efficiency and effectiveness.

## H. Extended Metrics

According to Tab. 1, the datasets are highly imbalanced. For instance, TCGA-BRCA and TCGA-NSCLC are dominated by invasive ductal carcinoma (IDC) and squamous cell carcinoma (SCC), respectively. Therefore, additional results of weighted and macro F1, as well as precision, recall, and AUC per class, should be investigated. We consider three evaluation scenarios: forward and reverse sequences of datasets in the **in-domain setting**, and the forward sequence in the **out-of-domain setting**. These results are reported in Tabs H.1, H.2, and H.3, respectively. All of these results are reported using a buffer size of 30 WSIs for all baseline continual learning methods. As a result, there are no consistent winners across methods in terms of class-wise precision, recall, and AUC. However, our MergeSlide consistently achieves the highest values among all methods. In particular, for rare classes, MergeSlide tends to outperform others in terms of precision and/or recall. For example, in Tab. H.1 (forward sequence, in-domain setting), considering the IDC of TCGA-BRCA, MergeSlide outperforms all other methods in both precision ($\geq$0.066) and AUC ($\geq$0.008). In the more challenging out-of-domain setting, where test slides are not acquired from the same sites as the training slides, this observation remains unchanged. MergeSlide maintains the highest precision and AUC for ILC, with values $\geq$0.034 and $\geq$0.013, respectively. A similar trend is observed for the adenocarcinoma class of TCGA-NSCLC, where MergeSlide outperforms all methods in precision, recall, and AUC, except for the precision in the out-of-domain setting. Moreover, for the ChRCC class of TCGA-RCC, MergeSlide consistently achieves the highest recall and AUC across all settings.

## I. Additional Analyses

### I.1. Time complexity

Regarding the formal time complexity of the orthogonal merging strategy, it consists of six steps (lines 15–20) in Algorithm 1 in the Supplementary: (1) computing the task vector $\Delta\tilde{\theta}_{1:t-1}$ between $\tilde{\theta}_{1:t-1}$ and $\theta_{base}$, (2) computing the task vector $\Delta\theta_t$ between $\theta_t$ and $\theta_{base}$, (3) performing SVD on $\tilde{\theta}_{1:t}$, (4) computing the projection $G(\Delta\theta_t)$, (5) computing $\lambda_t$, and (6) computing $\theta_t$. Given each set of parameters $\theta \in \mathbb{R}^{m \times n}$, the time complexity of steps (1), (2), (5), and (6) is $\mathcal{O}(m \cdot n)$, as they involve only addition, subtraction, or scalar–matrix operations. (3) involves SVD decomposition, for which we use a QR-based algorithm (see Algorithm 1a in [5]). The time complexity, in the case $m > n$, is $\mathcal{O}(m \cdot n^2)$. The projection in (4) is dominated by matrix multiplications with $U$ and $V^\intercal$, which also has a time complexity of $\mathcal{O}(m \cdot n^2)$. Obviously, these six steps are dominated by (3) and (4), then the time complexity for all steps is $\mathcal{O}(m \cdot n^2)$. Because the slide aggregator $f_\mathcal{A}$ includes $L$ layers, each layer has set of paramters $\theta_l \in \mathbb{R}^{m_l \times n_l}$, the total cost for orthogonal merging over all $L$ layers is $\mathcal{O}(\sum_{l=1}^{L} m_l n_l^2)$. For $|\mathcal{D}|$ datasets/tasks, the overall time complexity becomes $\mathcal{O}\left(|\mathcal{D}| \sum_{l=1}^{L} m_l n_l^2\right)$.

| Metric | Method | Subtype | MergeSlide (TCP) | MergeSlide (naive) | ADaFGrad | ConSlide | DER++ | AGEM | ER-ACE | Joint | Naive Finetuning |
|---|---|---|---|---|---|---|---|---|---|---|---|
| Balanced ACC (%) | | | **87.929** (±**2.110**) | 80.668 (±1.860) | 69.034 (±3.861) | 64.622 (±1.755) | 55.560 (±2.746) | 42.076 (±4.287) | 58.556 (±5.654) | 81.309 (±3.064) | 25.673 (±2.571) |
| Masked Balanced ACC (%) | | | **92.087** (±**1.740**) | **92.087** (±**1.740**) | 89.704 (±2.082) | 88.346 (±1.977) | 86.863 (±2.810) | 80.364 (±2.206) | 75.255 (±5.161) | 89.688 (±1.687) | 86.597 (±1.933) |
| Weighted F1 | | | **0.899** (±**0.013**) | 0.790 (±0.021) | 0.725 (±0.029) | 0.684 (±0.015) | 0.629 (±0.015) | 0.488 (±0.042) | 0.641 (±0.043) | 0.849 (±0.020) | 0.272 (±0.027) |
| Macro F1 | | | **0.860** (±**0.019**) | 0.793 (±0.020) | 0.688 (±0.037) | 0.644 (±0.015) | 0.544 (±0.026) | 0.398 (±0.039) | 0.561 (±0.058) | 0.827 (±0.024) | 0.222 (±0.037) |
| Precision per class | TCGA-BRCA | IDC | **0.947** (±**0.033**) | 0.944 (±0.035) | 0.960 (±0.039) | 0.969 (±0.031) | 0.881 (±0.064) | 0.703 (±0.236) | 0.820 (±0.079) | 0.918 (±0.041) | 0.000 (±0.000) |
| | | ILC | 0.730 (±0.133) | **0.745** (±**0.102**) | 0.684 (±0.153) | 0.613 (±0.104) | 0.672 (±0.291) | 0.018 (±0.054) | 0.401 (±0.418) | 0.816 (±0.132) | 0.633 (±0.425) |
| | TCGA-RCC | CC | **0.997** (±**0.008**) | 0.968 (±0.027) | 0.983 (±0.016) | 0.918 (±0.054) | 0.961 (±0.017) | 0.933 (±0.078) | 0.915 (±0.055) | 0.973 (±0.018) | 0.990 (±0.021) |
| | | P | 0.879 (±0.043) | 0.895 (±0.042) | 0.927 (±0.048) | 0.926 (±0.056) | **0.936** (±**0.040**) | 0.925 (±0.126) | 0.835 (±0.105) | 0.913 (±0.044) | 0.958 (±0.044) |
| | | ChRCC | 0.576 (±0.061) | 0.809 (±0.081) | 0.834 (±0.108) | 0.973 (±0.054) | **0.936** (±**0.088**) | 0.779 (±0.227) | 0.849 (±0.291) | 0.921 (±0.066) | 0.580 (±0.477) |
| | TCGA-NSCLC | SCC | **0.957** (±**0.029**) | 0.952 (±0.038) | 0.823 (±0.045) | 0.805 (±0.076) | 0.842 (±0.059) | 0.827 (±0.055) | 0.579 (±0.115) | 0.868 (±0.053) | 0.719 (±0.147) |
| | | A | **0.879** (±**0.029**) | 0.848 (±0.034) | 0.824 (±0.080) | 0.764 (±0.063) | 0.821 (±0.056) | 0.514 (±0.197) | 0.830 (±0.084) | 0.904 (±0.060) | 0.232 (±0.223) |
| | TCGA-ESCA | SCC | **0.943** (±**0.026**) | 0.748 (±0.063) | 0.805 (±0.281) | 0.761 (±0.281) | 0.400 (±0.490) | 0.597 (±0.421) | 0.529 (±0.234) | 0.803 (±0.066) | 0.367 (±0.458) |
| | | A | **0.907** (±**0.037**) | 0.410 (±0.035) | 0.477 (±0.414) | 0.280 (±0.368) | 0.000 (±0.000) | 0.000 (±0.000) | 0.427 (±0.312) | 0.707 (±0.118) | 0.000 (±0.000) |
| | TCGA-TGCT | S | 0.905 (±0.033) | 0.852 (±0.058) | 0.890 (±0.074) | **0.909** (±**0.085**) | 0.000 (±0.000) | 0.386 (±0.403) | 0.545 (±0.197) | 0.871 (±0.064) | 0.030 (±0.090) |
| | | MGCT | **0.761** (±**0.081**) | 0.749 (±0.147) | 0.623 (±0.346) | 0.721 (±0.324) | 0.100 (±0.300) | 0.292 (±0.375) | 0.307 (±0.415) | 0.843 (±0.153) | 0.000 (±0.000) |
| | TCGA-CESC | SCC | **0.649** (±**0.049**) | 0.594 (±0.075) | 0.458 (±0.066) | 0.414 (±0.063) | 0.338 (±0.064) | 0.308 (±0.063) | 0.026 (±0.077) | 0.777 (±0.092) | 0.133 (±0.040) |
| | | A | **0.959** (±**0.011**) | 0.824 (±0.039) | 0.634 (±0.040) | 0.598 (±0.019) | 0.555 (±0.025) | 0.453 (±0.037) | 0.697 (±0.080) | 0.821 (±0.051) | 0.372 (±0.026) |
| Recall per class | TCGA-BRCA | IDC | **0.938** (±**0.029**) | 0.918 (±0.027) | 0.787 (±0.052) | 0.677 (±0.051) | 0.875 (±0.063) | 0.786 (±0.264) | 0.879 (±0.076) | 0.934 (±0.044) | 0.000 (±0.000) |
| | | ILC | 0.746 (±0.183) | **0.875** (±**0.133**) | 0.882 (±0.131) | 0.917 (±0.087) | 0.839 (±0.301) | 0.100 (±0.300) | 0.318 (±0.366) | 0.775 (±0.191) | 0.176 (±0.188) |
| | TCGA-RCC | CC | 0.821 (±0.044) | 0.902 (±0.035) | 0.913 (±0.047) | 0.897 (±0.051) | **0.929** (±**0.032**) | 0.647 (±0.256) | 0.925 (±0.034) | 0.951 (±0.024) | 0.301 (±0.156) |
| | | P | 0.940 (±0.038) | **0.974** (±**0.024**) | 0.910 (±0.044) | 0.906 (±0.050) | 0.875 (±0.089) | 0.451 (±0.194) | 0.849 (±0.080) | 0.955 (±0.055) | 0.341 (±0.156) |
| | | ChRCC | **1.000** (±**0.000**) | 0.963 (±0.046) | 0.963 (±0.084) | 0.645 (±0.258) | 0.870 (±0.117) | 0.681 (±0.293) | 0.625 (±0.304) | 0.935 (±0.059) | 0.165 (±0.177) |
| | TCGA-NSCLC | SCC | 0.861 (±0.041) | 0.878 (±0.052) | 0.889 (±0.109) | 0.911 (±0.033) | 0.877 (±0.051) | 0.761 (±0.122) | **0.976** (±**0.024**) | 0.922 (±0.044) | 0.506 (±0.216) |
| | | A | **0.960** (±**0.026**) | 0.911 (±0.044) | 0.853 (±0.055) | 0.849 (±0.072) | 0.881 (±0.081) | 0.297 (±0.244) | 0.687 (±0.149) | 0.871 (±0.056) | 0.050 (±0.053) |
| | TCGA-ESCA | SCC | **0.869** (±**0.055**) | 0.859 (±0.047) | 0.193 (±0.091) | 0.176 (±0.122) | 0.017 (±0.023) | 0.052 (±0.047) | 0.500 (±0.332) | 0.752 (±0.100) | 0.031 (±0.042) |
| | | A | **0.961** (±**0.018**) | 0.787 (±0.049) | 0.050 (±0.059) | 0.018 (±0.031) | 0.000 (±0.000) | 0.000 (±0.000) | 0.200 (±0.244) | 0.532 (±0.163) | 0.000 (±0.000) |
| | TCGA-TGCT | S | 0.891 (±0.043) | 0.887 (±0.074) | 0.729 (±0.148) | 0.516 (±0.157) | 0.000 (±0.000) | 0.056 (±0.071) | **0.902** (±**0.083**) | 0.940 (±0.065) | 0.011 (±0.033) |
| | | MGCT | **0.784** (±**0.075**) | 0.585 (±0.142) | 0.327 (±0.252) | 0.313 (±0.171) | 0.009 (±0.027) | 0.062 (±0.086) | 0.190 (±0.285) | 0.587 (±0.224) | 0.000 (±0.000) |
| | TCGA-CESC | SCC | **0.790** (±**0.054**) | 0.575 (±0.087) | 0.755 (±0.088) | 0.785 (±0.092) | 0.770 (±0.078) | 0.785 (±0.098) | 0.090 (±0.270) | 0.660 (±0.109) | 0.800 (±0.063) |
| | | A | 0.913 (±0.020) | 0.514 (±0.080) | 0.966 (±0.021) | 0.962 (±0.016) | **0.966** (±**0.016**) | 0.966 (±0.017) | 0.685 (±0.241) | 0.891 (±0.042) | 0.968 (±0.018) |
| AUC per class | TCGA-BRCA | IDC | 0.940 (±0.032) | **0.951** (±**0.035**) | 0.908 (±0.052) | 0.912 (±0.047) | 0.858 (±0.049) | 0.627 (±0.140) | 0.789 (±0.103) | 0.935 (±0.044) | 0.393 (±0.134) |
| | | ILC | 0.945 (±0.030) | **0.965** (±**0.032**) | 0.948 (±0.044) | 0.957 (±0.029) | 0.944 (±0.032) | 0.757 (±0.100) | 0.895 (±0.074) | 0.957 (±0.037) | 0.915 (±0.057) |
| | TCGA-RCC | CC | 0.974 (±0.009) | **0.988** (±**0.008**) | 0.984 (±0.009) | 0.962 (±0.019) | 0.980 (±0.015) | 0.926 (±0.028) | 0.971 (±0.015) | 0.991 (±0.007) | 0.865 (±0.083) |
| | | P | 0.983 (±0.011) | **0.990** (±**0.007**) | 0.987 (±0.009) | 0.981 (±0.015) | 0.981 (±0.012) | 0.949 (±0.049) | 0.968 (±0.022) | 0.989 (±0.009) | 0.938 (±0.062) |
| | | ChRCC | 0.979 (±0.010) | 0.995 (±0.003) | 0.994 (±0.005) | **0.996** (±**0.004**) | 0.995 (±0.004) | 0.977 (±0.015) | 0.992 (±0.008) | 0.995 (±0.004) | 0.932 (±0.072) |
| | TCGA-NSCLC | SCC | **0.975** (±**0.015**) | 0.965 (±0.026) | 0.969 (±0.016) | 0.972 (±0.017) | 0.969 (±0.013) | 0.952 (±0.022) | 0.961 (±0.019) | 0.972 (±0.015) | 0.935 (±0.043) |
| | | A | **0.975** (±**0.015**) | 0.952 (±0.023) | 0.954 (±0.031) | 0.948 (±0.024) | 0.957 (±0.021) | 0.715 (±0.196) | 0.966 (±0.013) | 0.974 (±0.010) | 0.454 (±0.190) |
| | TCGA-ESCA | SCC | **0.977** (±**0.012**) | 0.971 (±0.012) | 0.841 (±0.076) | 0.575 (±0.149) | 0.576 (±0.090) | 0.701 (±0.114) | 0.838 (±0.096) | 0.886 (±0.046) | 0.705 (±0.049) |
| | | A | **0.977** (±**0.012**) | 0.967 (±0.012) | 0.914 (±0.022) | 0.904 (±0.036) | 0.924 (±0.020) | 0.927 (±0.022) | 0.918 (±0.034) | 0.941 (±0.020) | 0.918 (±0.031) |
| | TCGA-TGCT | S | 0.875 (±0.054) | 0.914 (±0.048) | 0.911 (±0.061) | 0.907 (±0.073) | 0.935 (±0.051) | **0.941** (±**0.055**) | 0.791 (±0.113) | 0.929 (±0.045) | 0.930 (±0.056) |
| | | MGCT | 0.884 (±0.053) | 0.802 (±0.091) | 0.817 (±0.093) | 0.790 (±0.128) | 0.812 (±0.123) | 0.822 (±0.112) | **0.925** (±**0.049**) | 0.934 (±0.051) | 0.691 (±0.226) |
| | TCGA-CESC | SCC | 0.952 (±0.016) | 0.915 (±0.041) | 0.969 (±0.010) | 0.958 (±0.018) | 0.968 (±0.010) | **0.970** (±**0.010**) | 0.363 (±0.263) | 0.966 (±0.017) | 0.962 (±0.014) |
| | | A | 0.951 (±0.017) | 0.878 (±0.031) | 0.946 (±0.017) | 0.953 (±0.019) | 0.964 (±0.015) | **0.967** (±**0.011**) | 0.866 (±0.184) | 0.944 (±0.019) | 0.960 (±0.017) |

Table H.1. Extended evaluation metrics in the sequence of B→R→N→E→T→C in the **in-domain setting**.

| Metric | Method | Subtype | MergeSlide (TCP) | MergeSlide (naive) | ADaFGrad | ConSlide | DER++ | AGEM | ER-ACE | Joint | Naive Finetuning |
|---|---|---|---|---|---|---|---|---|---|---|---|
| Balanced ACC (%) | | | **85.112** (±**1.570**) | 77.785 (±3.940) | 61.836 (±5.562) | 61.336 (±5.791) | 57.951 (±3.115) | 41.864 (±4.186) | 60.808 (±4.769) | 76.101 (±3.098) | 23.259 (±5.204) |
| Masked Balanced ACC (%) | | | **89.575** (±**2.440**) | **89.575** (±**2.440**) | 86.110 (±3.217) | 85.945 (±4.393) | 83.469 (±2.902) | 76.207 (±3.793) | 77.125 (±4.800) | 86.716 (±3.018) | 82.06 (±3.905) |
| Weighted F1 | | | **0.897** (±**0.008**) | 0.853 (±0.010) | 0.768 (±0.026) | 0.750 (±0.063) | 0.769 (±0.052) | 0.473 (±0.042) | 0.739 (±0.029) | 0.851 (±0.028) | 0.198 (±0.080) |
| Macro F1 | | | **0.830** (±**0.012**) | 0.741 (±0.012) | 0.600 (±0.071) | 0.587 (±0.071) | 0.569 (±0.040) | 0.341 (±0.059) | 0.555 (±0.044) | 0.753 (±0.026) | 0.156 (±0.068) |
| Precision per class | TCGA-BRCA | IDC | 0.941 (±0.015) | 0.941 (±0.029) | 0.942 (±0.044) | **0.964** (±**0.016**) | 0.898 (±0.062) | 0.467 (±0.382) | 0.876 (±0.067) | 0.907 (±0.054) | 0.000 (±0.000) |
| | | ILC | 0.714 (±0.087) | **0.759** (±**0.086**) | 0.677 (±0.201) | 0.607 (±0.109) | 0.712 (±0.170) | 0.192 (±0.293) | 0.725 (±0.259) | 0.768 (±0.099) | 0.488 (±0.489) |
| | TCGA-RCC | CC | **0.997** (±**0.005**) | 0.977 (±0.012) | 0.928 (±0.040) | 0.908 (±0.039) | 0.917 (±0.044) | 0.892 (±0.096) | 0.924 (±0.051) | 0.935 (±0.050) | 0.786 (±0.393) |
| | | P | 0.868 (±0.024) | 0.847 (±0.045) | **0.918** (±**0.039**) | 0.909 (±0.040) | 0.902 (±0.062) | 0.736 (±0.317) | 0.829 (±0.138) | 0.903 (±0.071) | 0.546 (±0.451) |
| | | ChRCC | 0.607 (±0.108) | 0.788 (±0.126) | 0.796 (±0.113) | 0.879 (±0.125) | 0.841 (±0.138) | **0.897** (±**0.207**) | 0.756 (±0.284) | 0.819 (±0.141) | 0.974 (±0.048) |
| | TCGA-NSCLC | SCC | **0.946** (±**0.047**) | 0.929 (±0.059) | 0.810 (±0.106) | 0.794 (±0.107) | 0.831 (±0.099) | 0.719 (±0.155) | 0.699 (±0.136) | 0.859 (±0.088) | 0.733 (±0.209) |
| | | A | 0.842 (±0.065) | **0.856** (±**0.055**) | 0.772 (±0.125) | 0.741 (±0.120) | 0.769 (±0.086) | 0.517 (±0.218) | 0.856 (±0.060) | 0.821 (±0.119) | 0.333 (±0.166) |
| | TCGA-ESCA | SCC | **0.967** (±**0.022**) | 0.651 (±0.217) | 0.531 (±0.443) | 0.564 (±0.395) | 0.408 (±0.429) | 0.343 (±0.433) | 0.348 (±0.336) | 0.652 (±0.268) | 0.200 (±0.400) |
| | | A | **0.891** (±**0.055**) | 0.385 (±0.203) | 0.217 (±0.350) | 0.200 (±0.400) | 0.125 (±0.301) | 0.000 (±0.000) | 0.262 (±0.209) | 0.584 (±0.229) | 0.000 (±0.000) |
| | TCGA-TGCT | S | **0.899** (±**0.030**) | 0.737 (±0.096) | 0.749 (±0.291) | 0.802 (±0.126) | 0.604 (±0.406) | 0.562 (±0.394) | 0.455 (±0.170) | 0.811 (±0.083) | 0.000 (±0.000) |
| | | MGCT | 0.443 (±0.356) | **0.455** (±**0.390**) | 0.375 (±0.437) | 0.451 (±0.425) | 0.175 (±0.354) | 0.100 (±0.300) | 0.056 (±0.096) | 0.698 (±0.322) | 0.000 (±0.000) |
| | TCGA-CESC | SCC | **0.652** (±**0.094**) | 0.469 (±0.154) | 0.395 (±0.174) | 0.352 (±0.178) | 0.401 (±0.215) | 0.221 (±0.120) | 0.119 (±0.299) | 0.726 (±0.135) | 0.039 (±0.019) |
| | | A | **0.962** (±**0.014**) | 0.741 (±0.062) | 0.513 (±0.172) | 0.518 (±0.164) | 0.526 (±0.169) | 0.309 (±0.080) | 0.708 (±0.184) | 0.756 (±0.120) | 0.172 (±0.062) |
| Recall per class | TCGA-BRCA | IDC | **0.930** (±**0.015**) | 0.904 (±0.022) | 0.779 (±0.091) | 0.675 (±0.086) | 0.838 (±0.081) | 0.511 (±0.374) | 0.851 (±0.057) | 0.907 (±0.043) | 0.000 (±0.000) |
| | | ILC | 0.748 (±0.061) | 0.829 (±0.054) | 0.825 (±0.100) | **0.890** (±**0.042**) | 0.626 (±0.208) | 0.390 (±0.471) | 0.582 (±0.279) | 0.704 (±0.175) | 0.083 (±0.116) |
| | TCGA-RCC | CC | 0.809 (±0.055) | 0.884 (±0.075) | 0.885 (±0.045) | 0.883 (±0.043) | **0.908** (±**0.052**) | 0.768 (±0.188) | 0.855 (±0.199) | 0.928 (±0.042) | 0.221 (±0.193) |
| | | P | 0.945 (±0.022) | **0.951** (±**0.024**) | 0.861 (±0.065) | 0.882 (±0.038) | 0.806 (±0.162) | 0.370 (±0.367) | 0.884 (±0.046) | 0.902 (±0.070) | 0.211 (±0.232) |
| | | ChRCC | **0.993** (±**0.015**) | 0.926 (±0.075) | 0.817 (±0.092) | 0.605 (±0.242) | 0.712 (±0.143) | 0.436 (±0.221) | 0.671 (±0.255) | 0.830 (±0.113) | 0.224 (±0.140) |
| | TCGA-NSCLC | SCC | 0.846 (±0.021) | 0.838 (±0.039) | 0.857 (±0.076) | 0.869 (±0.070) | 0.853 (±0.050) | 0.731 (±0.247) | **0.951** (±**0.023**) | 0.856 (±0.092) | 0.440 (±0.277) |
| | | A | **0.959** (±**0.021**) | 0.907 (±0.032) | 0.803 (±0.095) | 0.799 (±0.067) | 0.845 (±0.049) | 0.317 (±0.356) | 0.712 (±0.124) | 0.881 (±0.051) | 0.055 (±0.036) |
| | TCGA-ESCA | SCC | **0.831** (±**0.070**) | 0.817 (±0.049) | 0.144 (±0.176) | 0.069 (±0.062) | 0.091 (±0.097) | 0.034 (±0.043) | 0.357 (±0.394) | 0.689 (±0.182) | 0.008 (±0.017) |
| | | A | **0.973** (±**0.019**) | 0.803 (±0.081) | 0.026 (±0.060) | 0.012 (±0.030) | 0.008 (±0.016) | 0.000 (±0.000) | 0.488 (±0.362) | 0.517 (±0.241) | 0.000 (±0.000) |
| | TCGA-TGCT | S | 0.845 (±0.094) | **0.911** (±**0.139**) | 0.306 (±0.339) | 0.498 (±0.329) | 0.328 (±0.267) | 0.168 (±0.251) | 0.863 (±0.111) | 0.955 (±0.053) | 0.000 (±0.000) |
| | | MGCT | **0.536** (±**0.222**) | 0.440 (±0.415) | 0.254 (±0.333) | 0.232 (±0.224) | 0.020 (±0.041) | 0.050 (±0.150) | 0.350 (±0.391) | 0.476 (±0.217) | 0.000 (±0.000) |
| | TCGA-CESC | SCC | **0.810** (±**0.062**) | 0.547 (±0.098) | 0.757 (±0.079) | 0.784 (±0.128) | 0.763 (±0.107) | 0.808 (±0.053) | 0.052 (±0.121) | 0.540 (±0.208) | 0.800 (±0.111) |
| | | A | 0.904 (±0.053) | 0.497 (±0.073) | 0.962 (±0.027) | 0.951 (±0.037) | 0.965 (±0.023) | **0.966** (±**0.025**) | 0.486 (±0.276) | 0.834 (±0.135) | 0.968 (±0.024) |
| AUC per class | TCGA-BRCA | IDC | 0.941 (±0.014) | **0.946** (±**0.009**) | 0.891 (±0.039) | 0.915 (±0.036) | 0.847 (±0.041) | 0.677 (±0.196) | 0.822 (±0.105) | 0.912 (±0.027) | 0.468 (±0.098) |
| | | ILC | 0.945 (±0.013) | **0.965** (±**0.011**) | 0.949 (±0.016) | 0.952 (±0.022) | 0.927 (±0.027) | 0.801 (±0.121) | 0.919 (±0.045) | 0.951 (±0.015) | 0.888 (±0.034) |
| | TCGA-RCC | CC | 0.977 (±0.024) | **0.980** (±**0.019**) | 0.958 (±0.02) | 0.947 (±0.025) | 0.956 (±0.020) | 0.927 (±0.032) | 0.964 (±0.026) | 0.972 (±0.026) | 0.819 (±0.108) |
| | | P | 0.982 (±0.006) | **0.986** (±**0.005**) | 0.975 (±0.016) | 0.974 (±0.018) | 0.970 (±0.018) | 0.932 (±0.041) | 0.963 (±0.035) | 0.983 (±0.011) | 0.916 (±0.045) |
| | | ChRCC | 0.982 (±0.019) | **0.988** (±**0.013**) | 0.976 (±0.011) | 0.976 (±0.017) | 0.979 (±0.013) | 0.939 (±0.062) | 0.983 (±0.015) | 0.986 (±0.013) | 0.861 (±0.102) |
| | TCGA-NSCLC | SCC | **0.972** (±**0.005**) | 0.957 (±0.009) | 0.943 (±0.027) | 0.954 (±0.021) | 0.949 (±0.012) | 0.927 (±0.028) | 0.956 (±0.015) | 0.957 (±0.016) | 0.928 (±0.029) |
| | | A | **0.973** (±**0.004**) | 0.941 (±0.011) | 0.917 (±0.038) | 0.926 (±0.034) | 0.927 (±0.018) | 0.673 (±0.201) | 0.961 (±0.010) | 0.955 (±0.024) | 0.414 (±0.183) |
| | TCGA-ESCA | SCC | 0.980 (±0.020) | **0.989** (±**0.011**) | 0.494 (±0.127) | 0.556 (±0.112) | 0.414 (±0.209) | 0.565 (±0.176) | 0.839 (±0.228) | 0.863 (±0.058) | 0.647 (±0.196) |
| | | A | **0.980** (±**0.020**) | 0.943 (±0.056) | 0.930 (±0.053) | 0.912 (±0.036) | 0.911 (±0.047) | 0.939 (±0.032) | 0.921 (±0.072) | 0.956 (±0.033) | 0.943 (±0.027) |
| | TCGA-TGCT | S | 0.696 (±0.232) | 0.775 (±0.227) | 0.743 (±0.247) | 0.774 (±0.223) | 0.723 (±0.254) | **0.806** (±**0.199**) | 0.787 (±0.186) | 0.783 (±0.223) | 0.818 (±0.192) |
| | | MGCT | 0.704 (±0.225) | 0.607 (±0.354) | **0.777** (±**0.131**) | 0.777 (±0.205) | 0.505 (±0.182) | 0.657 (±0.180) | 0.814 (±0.191) | 0.838 (±0.113) | 0.681 (±0.169) |
| | TCGA-CESC | SCC | 0.959 (±0.013) | 0.926 (±0.032) | **0.967** (±**0.014**) | 0.962 (±0.015) | 0.959 (±0.024) | 0.964 (±0.023) | 0.465 (±0.243) | 0.972 (±0.017) | 0.958 (±0.021) |
| | | A | 0.955 (±0.014) | 0.895 (±0.034) | 0.958 (±0.010) | 0.956 (±0.017) | 0.955 (±0.018) | **0.964** (±**0.017**) | 0.826 (±0.098) | 0.928 (±0.022) | 0.960 (±0.013) |

Table H.2. Extended evaluation metrics on the sequence of B→R→N→E→T→C in the **out-of-domain setting**.

| Method / Metric | Subtype | MergeSlide (TCP) | MergeSlide (naive) | ADaFGrad | ConSlide | DER++ | AGEM | ER-ACE | Joint | Naive Finetuning |
|---|---|---|---|---|---|---|---|---|---|---|
| Balanced ACC (%) | | **87.930** (±2.112) | 80.636 (±1.865) | 77.096 (±4.474) | 70.821 (±2.180) | 61.037 (±3.319) | 43.961 (±3.009) | 54.613 (±4.323) | 81.151 (±2.283) | 24.976 (±3.33) |
| Balanced Masked ACC | | **92.109** (±1.700) | **92.109** (±1.700) | 90.216 (±1.677) | 88.896 (±2.442) | 84.758 (±1.929) | 79.458 (±3.215) | 79.428 (±3.682) | 90.138 (±1.854) | 86.614 (±3.543) |
| Weighted F1 | | **0.899** (±0.013) | 0.789 (±0.022) | 0.793 (±0.038) | 0.683 (±0.033) | 0.623 (±0.066) | 0.410 (±0.054) | 0.648 (±0.026) | 0.843 (±0.021) | 0.177 (±0.028) |
| Macro F1 | | **0.860** (±0.019) | 0.793 (±0.021) | 0.782 (±0.035) | 0.684 (±0.031) | 0.604 (±0.044) | 0.417 (±0.044) | 0.543 (±0.035) | 0.824 (±0.022) | 0.234 (±0.034) |
| Precision per class TCGA-BRCA | IDC | **0.947** (±0.033) | 0.944 (±0.035) | 0.719 (±0.129) | 0.433 (±0.045) | 0.433 (±0.448) | 0.000 (±0.000) | 0.000 (±0.000) | 0.768 (±0.137) | 0.000 (±0.000) |
| | ILC | 0.730 (±0.133) | 0.745 (±0.102) | 0.871 (±0.057) | **0.920** (±0.052) | 0.830 (±0.058) | 0.691 (±0.110) | 0.726 (±0.068) | 0.823 (±0.060) | 0.000 (±0.000) |
| TCGA-RCC | CC | **0.997** (±0.008) | 0.968 (±0.027) | 0.835 (±0.111) | 0.836 (±0.090) | 0.793 (±0.288) | 0.551 (±0.281) | 0.495 (±0.495) | 0.825 (±0.082) | 0.000 (±0.000) |
| | P | 0.879 (±0.043) | **0.898** (±0.040) | 0.748 (±0.146) | 0.667 (±0.166) | 0.422 (±0.331) | 0.058 (±0.126) | 0.200 (±0.400) | 0.883 (±0.120) | 0.000 (±0.000) |
| | ChRCC | 0.576 (±0.061) | 0.809 (±0.081) | 0.850 (±0.065) | 0.726 (±0.329) | **0.867** (±0.175) | 0.389 (±0.348) | 0.632 (±0.083) | 0.771 (±0.101) | 0.300 (±0.458) |
| TCGA-NSCLC | SCC | **0.957** (±0.029) | 0.952 (±0.038) | 0.675 (±0.108) | 0.415 (±0.063) | 0.408 (±0.170) | 0.109 (±0.134) | 0.456 (±0.088) | 0.740 (±0.094) | 0.150 (±0.320) |
| | A | 0.879 (±0.029) | 0.845 (±0.029) | 0.898 (±0.046) | **0.938** (±0.021) | 0.899 (±0.047) | 0.186 (±0.373) | 0.772 (±0.113) | 0.869 (±0.044) | 0.000 (±0.000) |
| TCGA-ESCA | SCC | **0.943** (±0.026) | 0.747 (±0.065) | 0.871 (±0.076) | 0.941 (±0.035) | 0.905 (±0.056) | 0.551 (±0.363) | 0.803 (±0.098) | 0.894 (±0.038) | 0.100 (±0.300) |
| | A | 0.907 (±0.037) | 0.412 (±0.038) | **0.971** (±0.028) | 0.954 (±0.047) | 0.935 (±0.057) | 0.916 (±0.060) | 0.704 (±0.096) | 0.968 (±0.025) | 0.910 (±0.075) |
| TCGA-TGCT | S | **0.905** (±0.033) | 0.852 (±0.058) | 0.870 (±0.040) | 0.873 (±0.108) | 0.780 (±0.104) | 0.328 (±0.084) | 0.736 (±0.108) | 0.928 (±0.043) | 0.729 (±0.188) |
| | MGCT | 0.761 (±0.081) | 0.743 (±0.151) | 0.806 (±0.108) | 0.666 (±0.062) | 0.853 (±0.163) | 0.786 (±0.166) | **1.000** (±0.000) | 0.914 (±0.099) | 0.840 (±0.159) |
| TCGA-CESC | SCC | 0.649 (±0.049) | 0.595 (±0.081) | 0.662 (±0.131) | 0.555 (±0.038) | 0.430 (±0.061) | 0.277 (±0.030) | **0.728** (±0.150) | 0.913 (±0.049) | 0.155 (±0.011) |
| | A | **0.959** (±0.011) | 0.822 (±0.041) | 0.808 (±0.081) | 0.640 (±0.158) | 0.707 (±0.151) | 0.697 (±0.169) | 0.416 (±0.428) | 0.821 (±0.120) | 0.644 (±0.139) |
| Recall per class TCGA-BRCA | IDC | **0.938** (±0.029) | 0.918 (±0.027) | 0.585 (±0.110) | 0.586 (±0.069) | 0.045 (±0.047) | 0.000 (±0.000) | 0.000 (±0.000) | 0.575 (±0.199) | 0.000 (±0.000) |
| | ILC | 0.746 (±0.183) | **0.875** (±0.133) | 0.740 (±0.114) | 0.448 (±0.133) | 0.456 (±0.222) | 0.344 (±0.273) | 0.770 (±0.070) | 0.896 (±0.051) | 0.000 (±0.000) |
| TCGA-RCC | CC | 0.821 (±0.044) | 0.902 (±0.035) | 0.848 (±0.182) | **0.915** (±0.048) | 0.462 (±0.406) | 0.679 (±0.345) | 0.125 (±0.220) | 0.962 (±0.024) | 0.000 (±0.000) |
| | P | 0.940 (±0.038) | **0.974** (±0.024) | 0.715 (±0.152) | 0.694 (±0.153) | 0.456 (±0.291) | 0.160 (±0.332) | 0.083 (±0.167) | 0.672 (±0.075) | 0.000 (±0.000) |
| | ChRCC | **1.000** (±0.000) | 0.963 (±0.046) | 0.641 (±0.129) | 0.069 (±0.045) | 0.183 (±0.105) | 0.331 (±0.319) | 0.655 (±0.140) | 0.710 (±0.162) | 0.010 (±0.016) |
| TCGA-NSCLC | SCC | 0.861 (±0.041) | **0.878** (±0.052) | 0.537 (±0.112) | 0.771 (±0.102) | 0.611 (±0.231) | 0.261 (±0.336) | 0.497 (±0.131) | 0.489 (±0.182) | 0.011 (±0.024) |
| | A | **0.960** (±0.026) | 0.909 (±0.043) | 0.835 (±0.056) | 0.741 (±0.045) | 0.796 (±0.063) | 0.099 (±0.199) | 0.911 (±0.071) | 0.918 (±0.028) | 0.000 (±0.000) |
| TCGA-ESCA | SCC | **0.869** (±0.055) | 0.862 (±0.046) | 0.812 (±0.105) | 0.693 (±0.069) | 0.734 (±0.052) | 0.116 (±0.120) | 0.789 (±0.091) | 0.882 (±0.062) | 0.002 (±0.007) |
| | A | 0.961 (±0.018) | 0.787 (±0.049) | 0.899 (±0.051) | 0.887 (±0.044) | 0.941 (±0.035) | 0.880 (±0.084) | **0.973** (±0.018) | 0.955 (±0.026) | 0.583 (±0.268) |
| TCGA-TGCT | S | 0.891 (±0.043) | 0.887 (±0.074) | 0.940 (±0.038) | **0.947** (±0.055) | 0.921 (±0.042) | 0.734 (±0.136) | 0.888 (±0.082) | 0.940 (±0.048) | 0.525 (±0.217) |
| | MGCT | 0.784 (±0.075) | 0.585 (±0.142) | 0.935 (±0.083) | **0.958** (±0.048) | 0.897 (±0.088) | 0.771 (±0.177) | 0.552 (±0.337) | 0.945 (±0.084) | 0.734 (±0.227) |
| TCGA-CESC | SCC | 0.790 (±0.054) | 0.570 (±0.090) | **0.972** (±0.020) | 0.963 (±0.019) | 0.960 (±0.023) | 0.951 (±0.027) | 0.850 (±0.082) | 0.938 (±0.041) | 0.938 (±0.072) |
| | A | **0.913** (±0.020) | 0.514 (±0.084) | 0.718 (±0.172) | 0.757 (±0.157) | 0.783 (±0.163) | 0.745 (±0.180) | 0.263 (±0.301) | 0.802 (±0.156) | 0.809 (±0.156) |
| AUC per class TCGA-BRCA | IDC | 0.940 (±0.032) | 0.951 (±0.034) | 0.944 (±0.037) | **0.953** (±0.017) | 0.535 (±0.132) | 0.641 (±0.148) | 0.914 (±0.052) | 0.963 (±0.019) | 0.298 (±0.102) |
| | ILC | 0.945 (±0.030) | **0.965** (±0.032) | 0.939 (±0.024) | 0.826 (±0.083) | 0.872 (±0.038) | 0.821 (±0.172) | 0.889 (±0.047) | 0.946 (±0.020) | 0.897 (±0.022) |
| TCGA-RCC | CC | 0.974 (±0.009) | **0.988** (±0.008) | 0.922 (±0.052) | 0.938 (±0.032) | 0.834 (±0.144) | 0.867 (±0.053) | 0.840 (±0.117) | 0.926 (±0.043) | 0.594 (±0.259) |
| | P | 0.983 (±0.011) | **0.990** (±0.007) | 0.942 (±0.052) | 0.948 (±0.051) | 0.780 (±0.129) | 0.864 (±0.072) | 0.681 (±0.193) | 0.924 (±0.051) | 0.791 (±0.162) |
| | ChRCC | 0.979 (±0.010) | **0.995** (±0.003) | 0.894 (±0.042) | 0.801 (±0.069) | 0.809 (±0.079) | 0.836 (±0.117) | 0.930 (±0.020) | 0.937 (±0.022) | 0.499 (±0.093) |
| TCGA-NSCLC | SCC | **0.975** (±0.015) | 0.965 (±0.020) | 0.850 (±0.050) | 0.907 (±0.031) | 0.819 (±0.083) | 0.677 (±0.215) | 0.778 (±0.092) | 0.885 (±0.047) | 0.700 (±0.089) |
| | A | **0.975** (±0.015) | 0.952 (±0.023) | 0.952 (±0.025) | 0.954 (±0.010) | 0.926 (±0.021) | 0.826 (±0.097) | 0.955 (±0.028) | 0.972 (±0.012) | 0.718 (±0.125) |
| TCGA-ESCA | SCC | **0.977** (±0.012) | 0.971 (±0.012) | 0.953 (±0.027) | 0.925 (±0.017) | 0.908 (±0.042) | 0.773 (±0.114) | 0.960 (±0.019) | 0.970 (±0.008) | 0.834 (±0.039) |
| | A | 0.977 (±0.012) | 0.967 (±0.012) | **0.985** (±0.009) | 0.985 (±0.011) | 0.982 (±0.012) | 0.970 (±0.021) | 0.980 (±0.012) | 0.991 (±0.005) | 0.904 (±0.077) |
| TCGA-TGCT | S | 0.875 (±0.054) | 0.914 (±0.048) | **0.988** (±0.007) | 0.988 (±0.010) | 0.985 (±0.011) | 0.970 (±0.014) | 0.983 (±0.007) | 0.991 (±0.006) | 0.949 (±0.025) |
| | MGCT | 0.884 (±0.053) | 0.801 (±0.092) | 0.994 (±0.006) | **0.996** (±0.004) | 0.995 (±0.004) | 0.976 (±0.028) | 0.991 (±0.013) | 0.997 (±0.004) | 0.967 (±0.042) |
| TCGA-CESC | SCC | 0.952 (±0.016) | 0.915 (±0.041) | 0.949 (±0.042) | **0.955** (±0.034) | 0.937 (±0.049) | 0.937 (±0.042) | 0.634 (±0.167) | 0.936 (±0.041) | 0.944 (±0.046) |
| | A | 0.951 (±0.017) | 0.878 (±0.031) | 0.958 (±0.035) | **0.958** (±0.032) | 0.943 (±0.043) | 0.947 (±0.036) | 0.798 (±0.161) | 0.956 (±0.040) | 0.946 (±0.039) |

Table H.3. Extended evaluation metrics on the reversed sequence of C→T→E→N→R→B in the **in-domain setting**.

## I.2. Performance Drop Comparison

When new tasks are added, the performance on previously learned tasks often degrades. We provide Fig. I.1 to examine the performance drop of each task as new tasks are sequentially introduced to the first five tasks. Overall, MergeSlide maintains the most stable performance across tasks. Notably, several tasks experience significant performance drops: TCGA-NSCLC, TCGA-ESCA, and TCGA-TGCT. For TCGA-NSCLC, DER++ shows a substantial drop after training on the last task (TCGA-CESC), while other methods appear more stable. For TCGA-ESCA, all models except MergeSlide show significant drops after training on both the fifth task (TCGA-TGCT) and the last task (TCGA-CESC). For TCGA-TGCT, although MergeSlide initially performs worse than others when the task is first introduced, it shows greater stability afterward. In contrast, other models suffer a noticeable decline in TCGA-TGCT performance after the final task is added.

## I.3. Confidence Score Study

To further explore model behavior in its predictions, we provide Fig. I.2, which shows the average confidence scores for the ground-truth cancer subtyping class positions. For instance, given the prediction logits $\hat{p}_i \in \mathbb{R}^C$ for each WSI, where $C$ is the total number of cancer subtypes so far, we take the $j$-th logit value $\hat{p}_{i,j}$ to visualize, where $j$ indicates the ground-truth cancer subtype. We observe the following: while other methods tend to show decreased confidence on recent tasks (e.g., ESCA, TGCT) and a significant shift toward the last task, CESC, MergeSlide maintains stable confidence across all six tasks, with no significant degradation on any of them.

## I.4. What Happens When the MLP is Also Trained?

As described, only the backbone $f_\mathcal{A}$ is trained when a new task is added, in order to obtain the corresponding weight set $\theta_t$. We do not train the MLP as a classifier but instead use class-aware prompt embeddings $E^{\mathcal{C},t}$. Results in Tab. I.1 show that the MLP-free strategy not only avoids performance degradation but also slightly improves both Masked ACC and Mean ACC compared to trainable MLPs, regardless of whether TCP is used. The only exception is overall ACC when using TCP, which exhibits a very marginal decrease.

| Method | MLP Setting | ACC | Masked ACC | Mean ACC |
|---|---|---|---|---|
| MergeSlide (naive) | Trainable MLP | 78.613 (±1.710) | 93.534 (±0.831) | 90.075 (±1.145) |
| | MLP-free (ours) | **78.797** (±1.916) | **93.617** (±1.098) | **90.640** (±1.247) |
| MergeSlide w/ TCP | Trainable MLP | **91.935** (±0.897) | 93.534 (±0.831) | 92.137 (±1.287) |
| | MLP-free (ours) | 91.914 (±0.980) | **93.617** (±1.098) | **92.686** (±0.899) |

Table I.1. Comparison between two strategies for MLP: Trainable MLP and MLP-Free on the sequence of B→R→N→E→T→C.

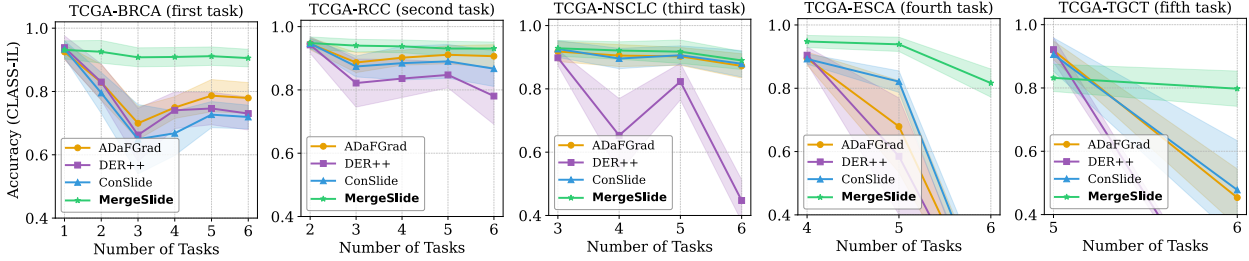

Figure I.1. Performance drop comparisons across different methods as new tasks are added to the first five datasets: TCGA-BRCA, TCGA-RCC, TCGA-NSCLC, TCGA-ESCA, and TCGA-TGCT (sequence: B→R→N→E→T→C).

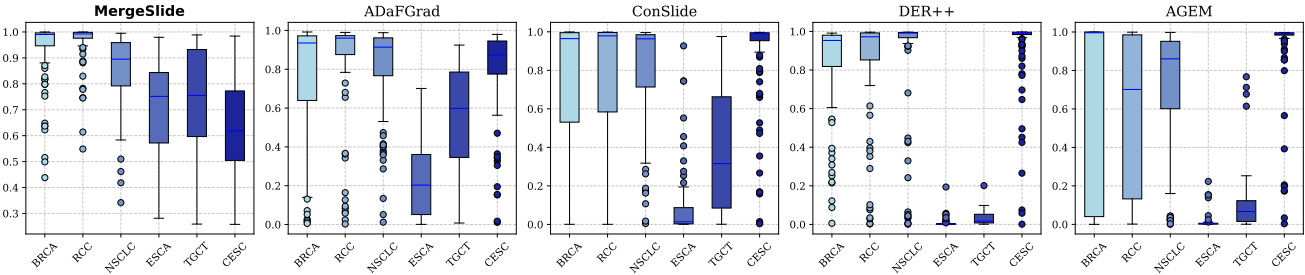

Figure I.2. Comparison of average confidence scores for target cancer subtyping after training on the last task, TCGA-CESC (sequence: B→R→N→E→T→C).

### I.5. Task-wise Performance

We provide a task-wise performance comparison between the top three methods under the CLASS-IL and TASK-IL scenarios in Tabs. I.3 and I.4, respectively, after training on the last task, i.e., TCGA-CESC. As a result, under the CLASS-IL scenario, MergeSlide still outperforms the others on five out of six tasks. Even without TCP, it achieves better performance than the others on four out of six tasks. However, we also observe that rehearsal-based methods tend to perform poorly on earlier tasks with limited WSIs, such as TCGA-ESCA and TCGA-TGCT. Indeed, after training on the final task (TCGA-CESC), the performance on this dataset becomes much higher compared to those with fewer samples. Nonetheless, MergeSlide maintains stable performance across tasks.

### I.6. Task Addition

Final accuracy after merging weights reflects the end performance; however, performance during the addition of new tasks should also be examined to ensure stability. The comparison of CLASS-IL ACC for MergeSlide against other methods is reported in Tab. I.2, where the number of accumulated tasks $T \in \{2, 3, 4, 5\}$ is evaluated. The results show that some methods exhibit unstable performance *while MergeSlide demonstrates consistent performances as the number of tasks increases.* $\sigma < 1\%$ further confirms the stability. We further compare performance drops on the two earliest tasks, TCGA-BRCA and TCGA-RCC, across top methods and MergeSlide without TCP (Fig. I.1). MergeSlide maintains stable performance as new tasks arrive, while others show significant and moderate drops on TCGA-BRCA and TCGA-RCC, respectively.

| Method | Number of increasing tasks | | | | |
|---|---|---|---|---|---|
| | $T=2$ | $T=3$ | $T=4$ | $T=5$ | $\sigma$ |
| ER-ACE | 90.045 ($\pm$1.968) | 81.210 ($\pm$2.793) | 79.297 ($\pm$4.234) | 74.849 ($\pm$5.461) | 5.525 |
| AGEM | 84.839 ($\pm$9.500) | 70.843 ($\pm$4.327) | 75.236 ($\pm$6.225) | 75.169 ($\pm$4.972) | 5.121 |
| DER++ | 92.244 ($\pm$3.177) | 86.693 ($\pm$2.215) | 88.104 ($\pm$1.323) | 82.307 ($\pm$2.946) | 3.549 |
| ConSlide | 86.733 ($\pm$2.228) | 80.643 ($\pm$3.226) | 83.929 ($\pm$1.968) | 85.257 ($\pm$2.068) | 2.250 |
| ADaFGrad | 90.945 ($\pm$3.214) | 86.002 ($\pm$2.225) | 87.382 ($\pm$1.908) | 85.252 ($\pm$2.271) | 2.187 |
| **MergeSlide** | **92.907** ($\pm$**1.337**) | **92.571** ($\pm$**0.713**) | **93.309** ($\pm$**0.493**) | **92.345** ($\pm$**0.990**) | **0.364** |

Table I.2. Closer look at MergeSlide compared to other continual learning methods as the number of tasks increases on the sequence of B→R→N→E→T→C (Metric: CLASS-IL ACC). $\sigma$ shows the standard deviation of results across $T \in \{2, 3, 4, 5\}$.

### I.7. Sensitivity of number of patches $K$

To further examine the sensitivity of the sampling patch number $K$, we report the bACC of MergeSlide under both CLASS-IL and TASK-IL settings for $K \in 50, 100, 200, 300, 400$ in Tab. I.3. In the CLASS-IL scenario, results without TCP remain stable, while with TCP the performance gap between $K = 50$ and $K = 400$ is only $4.997\%$. In the TASK-IL scenario, performance is also stable across different $K$ values. Notably, even the lowest setting ($K = 50$) still outperforms all comparative baselines.

| Method | BRCA | RCC | NSCLC |
|---|---|---|---|
| Zero-shot | 89.922 (±2.981) | 83.474 (±2.891) | 83.874 (±2.766) |
| DER++ [2] ($|\mathcal{B}_r|=30$) | 73.026 (±5.194) | 78.055 (±8.936) | 44.730 (±6.570) |
| ConSlide [7] ($|\mathcal{B}_r|=30$) | 71.940 (±3.949) | 86.700 (±5.921) | 87.933 (±4.409) |
| ADaFGrad [1] ($|\mathcal{B}_r|=30$) | 77.938 (±4.871) | 90.647 (±3.636) | 87.252 (±3.843) |
| **MergeSlide** | 90.784 (±3.311) | **93.418** (±2.351) | **93.657** (±2.633) |

| Method | ESCA | TGCT | CESC |
|---|---|---|---|
| Zero-shot | 62.388 (±3.704) | 79.771 (±5.316) | 17.578 (±2.928) |
| DER++ [2] ($|\mathcal{B}_r|=30$) | 2.836 (±2.707) | 0.526 (±1.053) | 93.516 (±1.310) |
| ConSlide [7] ($|\mathcal{B}_r|=30$) | 8.657 (±6.814) | 45.371 (±9.516) | **93.531** (±2.209) |
| ADaFGrad [1] ($|\mathcal{B}_r|=30$) | 6.269 (±5.247) | 47.787 (±15.559) | 93.594 (±1.148) |
| **MergeSlide** | **94.925** (±2.331) | **81.343** (±5.616) | 92.031 (±1.389) |

Figure I.3. Task-wise performances on stream of datasets B→R→N→E→T→C under CLASS-IL scenario.

| Method | BRCA | RCC | NSCLC |
|---|---|---|---|
| Zero-shot | 91.643 (±2.969) | 83.591 (±2.914) | 89.917 (±1.857) |
| DER++ [2] ($|\mathcal{B}_r|=30$) | 84.783 (±5.802) | 89.113 (±5.243) | 88.952 (±2.989) |
| ConSlide [7] ($|\mathcal{B}_r|=30$) | 87.604 (±4.920) | 92.574 (±4.466) | 92.265 (±2.503) |
| ADaFGrad [1] ($|\mathcal{B}_r|=30$) | 86.582 (±4.129) | **93.652** (±2.434) | 91.458 (±2.817) |
| **MergeSlide** | **93.359** (±3.353) | 93.645 (±2.100) | **92.939** (±2.539) |

| Method | ESCA | TGCT | CESC |
|---|---|---|---|
| Zero-shot | 88.060 (±2.111) | 81.862 (±5.855) | **95.859** (±1.108) |
| DER++ [2] ($|\mathcal{B}_r|=30$) | 87.185 (±3.379) | **92.153** (±3.806) | 93.913 (±1.201) |
| ConSlide [7] ($|\mathcal{B}_r|=30$) | 87.797 (±4.649) | 91.641 (±7.755) | 94.099 (±2.119) |
| ADaFGrad [1] ($|\mathcal{B}_r|=30$) | 88.152 (±3.832) | 90.799 (±7.266) | 94.362 (±1.164) |
| **MergeSlide** | **95.075** (±1.772) | 91.066 (±2.433) | 94.297 (±1.443) |

Figure I.4. Task-wise performances on stream of datasets B→R→N→E→T→C under TASK-IL scenario.

| $K$ | bACC (CLASS-IL) | | Masked bACC (TASK-IL) |
|---|---|---|---|
| | MergeSlide w/o TCP | MergeSlide w/ TCP | MergeSlide |
| 50 | 78.810 (±1.942) | 82.932 (±1.802) | 90.935 (±1.870) |
| 100 | 80.167 (±1.752) | 84.284 (±2.478) | 91.610 (±1.535) |
| 200 | 80.078 (±1.952) | 86.269 (±1.975) | 91.761 (±1.907) |
| 300 | 79.849 (±1.944) | 87.416 (±2.015) | 92.083 (±1.514) |
| 400 | **80.668** (±1.860) | **87.929** (±2.110) | **92.087** (±1.740) |

Table I.3. Sensitivity of $K$ sampling patches under the CLASS-IL and TASK-IL scenarios on B→R→N→E→T→C.



\end{document}